\documentclass[12pt]{article}

\usepackage[a4paper, total={6in, 8in}]{geometry}
\usepackage[T1]{fontenc}
\usepackage{lmodern}
\usepackage{microtype}

\usepackage{booktabs} 
\usepackage[english]{babel}
\usepackage[utf8x]{inputenc}
\usepackage[T1]{fontenc}
\usepackage{comment}
\usepackage[export]{adjustbox}
\usepackage{booktabs}

\usepackage{amsmath}
\usepackage{graphicx}
\usepackage{amsmath}
\usepackage{bm}
\usepackage{amsthm}
\usepackage{amssymb}
\usepackage{bbm}
\usepackage{verbatim}
\usepackage{graphicx}
\usepackage{afterpage}
\usepackage{etoolbox}
\usepackage{float}
\usepackage{rotating}
\usepackage[inline]{enumitem}
\usepackage{diagbox}

\usepackage{multirow}
\usepackage{caption}
\usepackage[skip=0pt]{subcaption}
\newdimen\figrasterwd
\figrasterwd\textwidth

\usepackage{algorithmic}
\usepackage{algorithm}
\newcommand {\myvec}[1] {{\mbox{\boldmath $#1$}}}

\newcommand{\myX}{\myvec{X}}

\newcommand{\myY}{\myvec{Y}}

\newcommand{\R}{\mathbb{R}}
\newcommand{\M}{\mathcal M}

\usepackage{hyperref}

\usepackage{titling}

\title{Multi-modal Differentiable Unsupervised Feature Selection}
\date{}

\author{ Junchen Yang $^{1}$ \and Ofir Lindenbaum $^{2}$ \and Yuval Kluger$^{1}$ \and Ariel Jaffe$^{3 \dagger}$\\
\normalsize{$^{1}$Yale University, USA;}\\
\normalsize{$^{2}$Bar-Ilan University, Israel;}\\
\normalsize{$^{3}$Hebrew University of Jerusalem, Israel}\\
\normalsize{$^\dagger$Corresponding author. E-mail: ariel.jaffe@mail.huji.ac.il}\\
}
\begin{document}
\maketitle

\begin{abstract}
Multi-modal high throughput biological data presents a great scientific opportunity and a significant computational challenge. In multi-modal measurements, every sample is observed simultaneously by two or more sets of sensors. In such settings, many observed variables in both modalities are often nuisance and do not carry information about the phenomenon of interest. Here, we propose a multi-modal unsupervised feature selection framework: identifying informative variables based on coupled high-dimensional measurements. Our method is designed to identify features associated with two types of latent low-dimensional structures: (i) shared structures that govern the observations in both modalities and (ii) differential structures that appear in only one modality. To that end,  we propose two Laplacian-based scoring operators. We incorporate the scores with differentiable gates that mask nuisance features and enhance the accuracy of the structure captured by the graph Laplacian. The performance of the new scheme is illustrated using synthetic and real datasets, including an extended biological application to single-cell multi-omics.     
\end{abstract}

\section{Introduction}\label{sec:intro}
In an effort to study biological systems, researchers are developing cutting-edge techniques that measure up to tens of thousands of variables at single-cell resolution. The complexity of such systems requires collecting multi-modal measurements to understand the interplay between different biological processes. Examples of such multi-modal measurements include SHARE-seq \cite{ma2020chromatin}, DBiT-seq \cite{liu2020high}, CITE-seq \cite{stoeckius2017simultaneous}, etc., which have provided biological insights and advancements in applications such as transcription factor characterization \cite{joung2023transcription}, cell type identification in human hippocampus \cite{xiao2022spatially}, and immune cell profiling \cite{leblay2020cite}.

Multi-modal learning is a powerful tool widely used across multiple disciplines to extract latent information from high-dimensional measurements \cite{sun2013survey,yan2021deep}. Humans use complementary senses when attempting to ``estimate'' spoken words or sentences \cite{raij2000audiovisual}. For example, lip movements can help us distinguish between two syllables that sound similar. The same intuition has inspired statisticians and machine learning researchers to develop learning techniques that exploit information captured simultaneously by complementary measurement devices.

Due to their applicability in multiple domains, there has been a growing interest in multi-modal approaches. Algorithms such as Contrastive Language–Image Pre-training
(CLIP) \cite{radford2021learning}, and Audioclip \cite{guzhov2022audioclip} have pushed the performance boundaries of machine learning for image, text, audio, analysis, and synthesis. The multi-modal data fusion task dates back to \cite{CCA1}, which proposed the celebrated Canonical Correlation Analysis (CCA). CCA has many extensions \cite{DCCA,lindenbaum2022lsparse}, and applications in diverse scientific domains \cite{cca_bio,cca_fault}. Despite their tremendous success, classical or advanced multi-modal schemes are often 
unsuitable for analyzing biological data. The large number of nuisance variables, which often exceeds the number of measurements, often causes correlation-based methods to overfit. 


To attenuate the influence of nuisance or noisy features, several authors proposed unsupervised feature selection (UFS) schemes \cite{solorio2020review}. 
UFS seeks small subsets of informative variables in order to improve downstream analysis tasks, such as clustering or manifold learning. Empirical results demonstrate that informative features are often smooth with respect to some latent structure \cite{degeest2018smoothness}. In practice, the smoothness of features can be evaluated based on how slowly they vary with respect to a graph \cite{he2005laplacian}. Follow-up works exploited this idea to identify informative features \cite{zhao2012spectral,shaham2021deep}. An alternative paradigm for UFS seeks subsets of features that can be used to reconstruct the entire data effectively \cite{balin2019concrete}.

While most fusion methods focus on extracting information shared between modalities, 
we propose a multi-modal UFS framework to identify 
features associated both with structures that appear in both modalities, and
structures that are \textit{modality-specific}, and appear in only one modality.
To capture the shared structure, we construct a symmetric shared graph Laplacian operator that enhances the shared geometry across modalities. We further propose differential graph operators that capture smooth structures that are not shared with the other modality. To perform multi-modal feature selection, we incorporate differentiable gates \cite{yamada2020feature,yang2022locally} with the \textit{shared} and \textit{modality-specific} graph Laplacian scoring functions. This leads to a differentiable UFS scheme that attenuates the influence of nuisance features during training and computes a more accurate Laplacian matrix \cite{lindenbaum2021differentiable}.

Our contributions are four folds: (i) Develop a \textit{shared} and \textit{modality-specific} Laplacian scoring operators. (ii) Motivate our operators using a product of manifolds model. (iii) develop and implement a differentiable framework for multi-modal UFS. (iv) Evaluate the merits and limitations of our approach with synthetic and real data and compare it to existing schemes.


\section{Problem setting and preliminaries}
\label{sec:prem_method}
We are given two data matrices 
$\myX \in 
\R^{n\times d},\myY \in \R^{n\times m}$ whose rows contain $n$ observations captured simultaneously in two modalities. The two sets of observations can be, for example, two arrays of sensors, cameras with different angles, etc. We are interested in processing modalities with bijective correspondences, which implies that there is a registration between the observations in both modalities. 

\begin{figure*}[htb!] 
    \centering
    
    \includegraphics[width=0.6\textwidth]{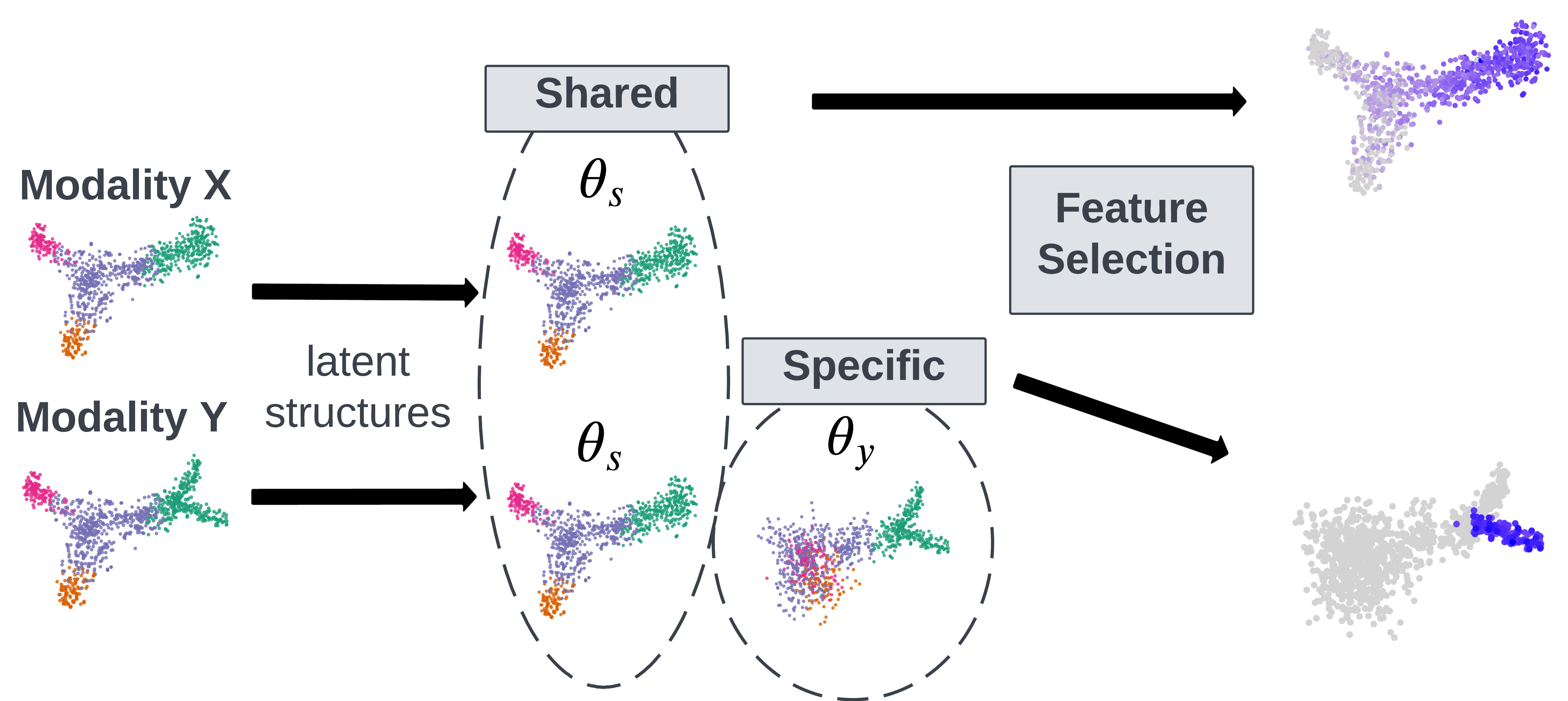}
    
    \caption{Overview of the goal: discovering features associated with shared and modality specific latent structures}%
    \label{fig:workflow}
\end{figure*}

Though the observations are high-dimensional, we assume that there are a small number of parameters governing the physical processes that underlies the data. These parameters can be continuous such as in a developmental process, or discrete - for example, when the observations can be characterized by clustering. 
However, the latent structure in both modalities may not be identical. For example, the two sets of observations may be generated by sets of sensors with different resolutions or sensitivity. 
For illustration, consider the observations shown in Fig. \ref{fig:workflow} (left). Both  modalities follow a very similar tree structure. 
The bottom tree, however, has an additional bifurcating point that does not appear in the upper tree (green points). 

Thus, we assume the latent parameters can be partitioned into two subsets. The first component denoted $\myvec{\theta}_s$, captures the structures shared by both modalities. 
The second component, denoted $\myvec{\theta}_x$ for modality $\myX$, and $\myvec{\theta}_y$ for modality $\myY$, captures the modality-specific structures that only appear in one set of observations. For example, the additional branch in the bottom tree (modality $\myY$) in Fig. \ref{fig:workflow} is governed by a parameter in $\myvec{\theta}_{y}$. 
Thus, the observations $\myX$ and $\myY$ are nonlinear transformations of $\myvec{\theta}_{s},\myvec{\theta}_{x}$ and $\myvec{\theta}_{s},\myvec{\theta}_{y}$, respectively.

Many biological data modalities are high dimensional and contain noisy features, which hinders the discovery of the underlying shared or modality-specific structures. Here, our goal is to identify groups of features  associated with the shared structures $\myvec{\theta}_{s}$ (e.g., the groups of features that are smooth on the shared bifurcated tree in Fig. \ref{fig:workflow}) and groups of features  associated with the modality-specific structures $\myvec{\theta}_{x}$ and $\myvec{\theta}_{y}$ (e.g., the features that are smooth with respect to the additional branch ($\myvec{\theta}_{y}$) of modality $\myvec{Y}$ in Fig. \ref{fig:workflow}). To achieve this goal, we compute two graphs that correspond to the two modalities. We use a spectral method to uncover the shared and graph-specific structures and apply a feature selection method to detect variables relevant to these structures. To better understand our approach, we  first introduce some preliminaries about graph representation in Sec. \ref{sec:graph_th}, and discuss related work on feature selection in Sec. \ref{sec:dufs}.



\subsection{The graph Laplacian and Laplacian score}
\label{sec:graph_th}

A common assumption when analyzing high-dimensional datasets is that their structure lies on a low dimensional manifold in the high dimensional space \cite{linderman2019fast,peterfreund2020local}. Methods for manifold learning are often based on a graph that captures the affinities between data points. Let $\myvec{x}^{(i)},\myvec{y}^{(i)} $ denote the $i$-th observation in the $\myX$ and $\myY$ modalities and let $\myvec{K}_x,  \myvec{K}_y$ be, respectively, their affinity matrices whose elements are computed by the following Gaussian kernel functions. 
\begin{align*}
    {(\myvec{K}_{x})}_{i,j} &= \exp{\big(-\frac{\|\myvec{x}^{(i)} - \myvec{x}^{(j)}\|^{2} }{2\sigma_x^2}\big)}, \\
    {(\myvec{K}_{y})}_{i,j} &= \exp{\big(-\frac{\|\myvec{y}^{(i)} - \myvec{y}^{(j)}\|^{2} }{2\sigma_y^2}\big)},
\end{align*}
where $\sigma_x,\sigma_y$ are user-defined bandwidths that control the decay of each Gaussian kernel. Intuitively, the affinities decay exponentially with the distances between samples, thus capturing the local neighborhood structure in the high-dimensional space.

We compute the normalized Laplacian matrix by $\myvec{L}_{x} = \myvec{D}^{-\frac{1}{2}}_{x}\myvec{K}_{x}\myvec{D}^{-\frac{1}{2}}_{x}$, where $\myvec{D}_{x}$ is a diagonal matrix of row sums of $\myvec{K}_x$. Similarly, $\myvec{L}_y$ is computed for modality $\myvec{Y}$. An important property of the Laplacian matrix is that its eigenvectors corresponding to large eigenvalues reflect the underlying geometry of the data. The Laplacian eigenvectors are used for  many applications including data embeddings \cite{belkin2003laplacian}, clustering \cite{von2007tutorial}, and feature selection \cite{he2005laplacian}. For the latter, a popular metric for unsupervised identification of informative features is the Laplacian Score (LS) \cite{he2005laplacian}, 
\begin{equation}
    \myvec{f}^{T}\myvec{L}_{x}\myvec{f} = \sum_{i=1}^{n}\lambda_{i} (\myvec{f}^T\myvec{u}_i)^2,
    \label{eq:LS}
\end{equation}
where $\myvec{L}_{x} = \sum_{i=1}^{n}\lambda_{i} \myvec{u}_i \myvec{u}_i^T$ is the eigendecomposition of $\myvec{L}_{x}$ and $\myvec{f}$ is the normalized feature vector. Intuitively, when $\myvec{f}$ varies slowly with respect to the underlying structure of $\myvec{L}_{x}$, it will have a significant component projected onto the subspace of its top eigenvectors, and a higher score. 

\subsection{Differentiable Unsupervised Feature Selection}
\label{sec:dufs}


A key limitation of the Laplacian score stems from its underlying assumption that the Laplacian matrix $\myvec{L}_{x}$ accurately reflects the latent structure of the data. This assumption, however,  may not be valid in the presence of many noisy features. In such cases, the top eigenvectors of $\myvec{L}_{x}$ may be heavily influenced by noise and would not capture the underlying structure accurately. A recent work \cite{lindenbaum2021differentiable} addresses this problem by developing Differentiable Unsupervised Feature Selection (DUFS), a framework that estimates the Laplacian matrix while simultaneously selecting informative features using Laplacian scores.
Specifically, DUFS computes a binary vector $\myvec{s} \in \{0,1\}^{d}$ that indicates which features are kept ($s_j = 1$) and which features are not ($s_j = 0$). Let $\Delta(\myvec{s})$ denote a diagonal matrix with $\myvec{s}$ on the diagonal. At each iteration of DUFS, the Laplacian is computed based on 
$\myvec{\tilde{X}} = \myvec{X}\Delta(\myvec{s})$, while simultaneously updaing $\myvec{s}$ by optimizing over the following loss function.  

\begin{equation}
     \mathcal{L} = -\frac{1}{n} \text{Tr}[\myvec{\tilde{X}}^T \myvec{L}_{\tilde{x}} \myvec{\tilde{X}}] + \lambda \|\myvec{s} \|_0,
    \label{eq:DUFS_loss1}
\end{equation}
where $\text{Tr[]}$ denotes the matrix trace.  The first term equals the sum of Laplacian Scores across all features normalized by the total number of samples $n$ in a training batch. The second term is a $\ell_0$ regularizer that imposes sparsity to the number of selected features, with $\lambda$ being a tunable parameter that controls the sparsity level. The output of DUFS is a list of a small number of selected features, and the Laplacian matrix $\myvec{L}_{\tilde{x}}$ learned from them. 

\label{sec:stg}
However, due to the discrete nature of the $\ell_0$ regularizer, the standard discrete indicator vector $\myvec{s} \in \{0,1\}^D$ will make objective in Eq. \eqref{eq:DUFS_loss1} not differentiable and finding the optimal solution intractable. Following, \cite{yamada2020feature}, one can relax the $\ell_0$ norm to a probabilistic differentiable counterpart, by replacing the binary indicator vector $\myvec{s}$ with a relaxed Bernoulli vector $\myvec{z}$. Specifically, $\myvec{z}$ is a continuous Gaussian reparametrization of the discrete random variables, termed Stochastic Gates.
It is defined for each feature $i$:
\begin{equation}
    z_i = \max(0,\min(1,0.5+ \mu_i + \epsilon_i)), \quad \epsilon_i \sim \mathcal{N}(0 ,\sigma^2)
\label{eq:z_stg}
\end{equation}
where $\mu_i$ is a learnable parameter, and $\sigma$ is fixed throughout training. 
The loss function in Eq. \eqref{eq:DUFS_loss1} can now be reformulated as follows, which is the final objective of the DUFS:
\begin{equation}
    \mathcal{L} = -\frac{1}{n} \text{Tr}[\myvec{\tilde{X}}^T \myvec{L}_{\tilde{x}} \myvec{\tilde{X}}] + \lambda \|\myvec{z} \|_0.
    \label{eq:DUFS_loss2}
\end{equation}

\section{Method}
\label{sec:method}
We now derive our approach for unsupervised feature selection in multi-modal settings. 
Our method is designed to capture two types of features: (i)  
 Features associated with latent structures that are \textit{shared} between two modalities.   
    (ii) Features associated with  \textit{differential 
 latent structures}, that appear in only one modality.
 In Sec. \ref{sec:joint_op} and \ref{sec:diff_op}, we derive two operators designed to capture shared and differential structures, respectively. 
To motivate our approach and illustrate the difference between shared and differential structures, we specifically address two examples: (i) shared and differential clusters and (ii) product of manifolds. We use the proposed operators in Sec. \ref{sec:mvdufs} to derive mmDUFS. 



\subsection{The shared structure operator}
\label{sec:joint_op}
\begin{figure*}[!htb] %
    \centering
    
    \includegraphics[width=0.9\textwidth]{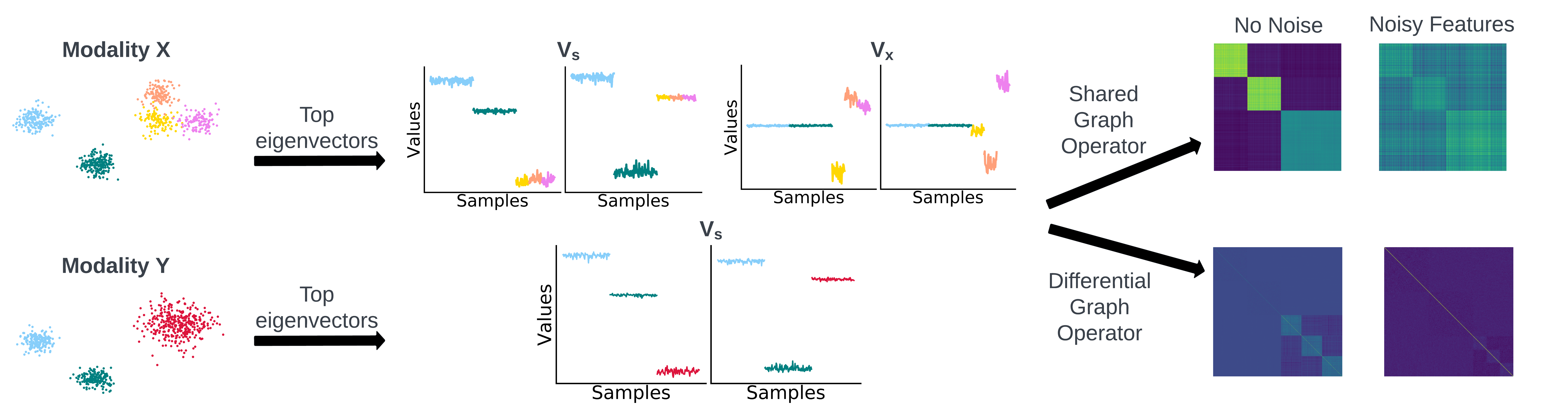}
    
    \caption{Visualization of the eigenvectors and the affinity matrix  of the proposed operators on an artificial cluster example. Left: Visualization of the clusters. Middle: Leading eigenvectors of $\myvec{L}_x$ and $\myvec{L}_y$.  Right: Affinity matrices of the proposed shared graph operator (top) and the differential graph operator (bottom) with/without the presence of noisy features.}
    \label{fig:vis_joint_op}
\end{figure*}

To motivate our approach,  
let us consider the artificial example illustrated in Fig. \ref{fig:vis_joint_op}. 
The lower figure in the left panel shows the observations in modality $\myY$, which contains samples from a mixture of three distinct Gaussians. The upper figure shows modality $\myX$, where one of the three clusters is partitioned again into three (less distinct) clusters.

It is instructive to study the \textit{ideal setting} where we make the following assumptions: (i) The largest distance between two nodes within a cluster, denoted $d_{\text{within}}$ is much smaller than the 
smallest distance between pairs of nodes of two clusters, denoted  $d_{\text{between}}$. (ii) The bandwidth $\sigma_x,\sigma_y$ is chosen such that $d_{\text{within}} \ll \sigma_x,\sigma_y \ll d_{\text{between}}$. In this setting, the three Gaussians constitute three main clusters, with no connections between pairs of nodes of different clusters and similar weights between pairs of nodes within clusters. 
Thus, the leading  eigenvectors of $\myvec{L}_y$ span the subspace of the three \textit{indicator vectors}. That is vectors that contain the square root of the degree of a node in a cluster and a zero value outside the cluster. See \cite{von2007tutorial} and illustration in 
Fig. \ref{fig:vis_joint_op}.  The matrix $\myvec{L}_x$ has two extra significant eigenvectors that span the separation of the third cluster, which appears only in $\myX$. 
We denote by $\myvec{V}_s$ a matrix that contains the indicator vectors of the three partitions that appear in $\myX$ and $\myY$ and by $\myvec{V}_x$ a matrix that contains the partitions that appear only in $\myX$.
In our ideal setting, the two Laplacian matrices $\myvec{L}_x,\myvec{L}_y$ are equal to
\begin{equation}
    \myvec{L}_x \approx \myvec{V}_s  \myvec{V}_{s}^{T} +  \myvec{V}_x  \myvec{V}_{x}^{T}, \qquad
    \myvec{L}_y \approx \myvec{V}_s  \myvec{V}_{s}^{T}. \label{eq:Laplacian_rewrite}
\end{equation}

To capture \textit{shared} latent structures we compute the following shared operator $\myvec{P}_{\text{shared}}$,
\begin{equation}
    \myvec{P}_{\text{shared}} = \myvec{L}_x \myvec{L}_y + \myvec{L}_y \myvec{L}_x.
    \label{eq:composite_op}
\end{equation}
For the cluster setting, the orthogonality between the matrices $\myvec{V}_s,\myvec{V}_x$ implies $
     \myvec{P}_{\text{shared}} \approx 2\myvec{V}_s \myvec{V}_{s}^{T}.
$

The symmetric product of the two Laplacians captures clusters that appear in both modalities while removing modality-specific clusters, see right panel of Fig. \ref{fig:vis_joint_op}.
We note that a similar operator to Eq. \eqref{eq:composite_op} is proposed in \cite{shnitzer2019recovering} for computing low-dimensional representations. Here, we combine our operator with DUFS to develop a multi-modal feature selection pipeline. 
We illustrate the usefulness of the shared operator for the product of manifold setting.  


\paragraph{Product of manifolds.}
Let $\M_a,\M_b$ and $\M_s$ be three
low-dimensional manifolds embedded in $\R^n$, which are smooth transformations of three sets of latent variables $\myvec{\theta}_a,\myvec{\theta}_b$ and $\myvec{\theta}_s$.
To further motivate our approach, consider the case where modalities $\myX$ and $\myY$ each contains observations from the products
$\M_y,\M_x$ given by,
\[
\M_y = \M_s \times \M_a, \qquad \M_x = \M_s \times \M_b.
\]
Note that the dependence on $\myvec{\theta}_s$ is shared between $\M_x, \M_y$, while the dependence on $\myvec{\theta}_a,\myvec{\theta}_b$ is modality-specific.  

In a product of manifolds $\M_x = \M_s \times \M_b$, every point $\myvec{x} \in \M_x$ is associated with two points $\myvec{x}_s \in \M_s$ and $\myvec{x}_b \in \M_b$. Thus, we can define projection operators $\pi^x_b(\myvec{x}),\pi^x_s(\myvec{x})$ that map a point $\myvec{x}$ in $\M_x$ to points in $\M_b,\M_s$, respectively. In addition, for every function $f^b: \M_b \to \R$ we define its extension to the product manifold $\M_x$ by
\[
(f^b \circ \pi_b^x)(\myvec{x})  = f^b( \pi^x_b(\myvec{x})).
\]
An important property of a product $\M_x$ is that the eigenfunctions $f_{l,m}^x$ of the Laplace Beltrami operator are  equal to the pointwise product of the eigenfunctions of  $\M_b,\M_s$, extended to $\M_x$. 
\begin{equation}\label{eq:eigenfunctions_product}
f^x_{l,m} = (f^s_l \circ \pi_s^x) (f^b_m \circ \pi_b^x). 
\end{equation}
We refer to \cite{zhang2021product} for a detailed description of the properties of the product of manifolds.
A simple example of a product of manifolds is a 2D rectangle area $(\myvec{\theta}_s,\myvec{\theta}_b) \in [0,l_s] \times [0,l_b]$. 
the projection $\pi_s^x$ yields the first coordinate, while $\pi_b^x$ yields the second. 
The eigenfunctions of the product with Neumann boundary conditions are equal to,
\begin{equation}\label{eq:rectangle}
f_{l,m} = \cos(\pi l \myvec{\theta}_s/l_s) \cos(\pi m \myvec{\theta}_b/l_b). 
\end{equation}

\paragraph{Observations generated uniformly at random over the product of manifolds.}
Here, we assume that the observations in the two modalities are generated by random and independent uniformly distributed samples over $\M_x,\M_y$. Let $\myvec{\phi}_{l,m}^x(\myvec{x}_i),\myvec{\phi}_{l,k}^y(\myvec{y}_i)$ denote the eigenvectors of $\myvec{L}_x,\myvec{L}_y$ evaluated at $\myvec{x}_i,\myvec{y}_i$ respectively. 
In the asymptotic regime where the number of points $n \to \infty$, the eigenvectors converge to the eigenfunctions as characterized in Eq. \eqref{eq:eigenfunctions_product}.
\begin{align}
\myvec{\phi}_{l,m}^x(\myvec{x}_i) &= \myvec{\phi}^s_l( \pi_s^x(\myvec{x}_i)) \myvec{\phi}^b_m(\pi_b^x(\myvec{x}_i)) \notag \\
\myvec{\phi}_{l,k}^y(\myvec{y}_i) &= \myvec{\phi}^s_l(\pi_s^y(\myvec{y}_i)) \myvec{\phi}^a_k(\pi_a^y(\myvec{y}_i)). 
\end{align}

Details about the definition and rate of convergence can be found, for example, in \cite{cheng2022eigen,garcia2020error}, and reference therein.
It is instructive to consider the ideal case, where due to their dependence on the independent projections $\pi^x_b$ and $\pi^x_a$, the eigenvectors $\myvec{\phi}_{l,m}^x,\myvec{\phi}_{l,k}^y$ satisfy the following 
orthogonality property,
\begin{equation}
(\myvec{\phi}^x_{l,m})^T \myvec{\phi}^y_{l',k}  
= 
\begin{cases}
1 & l = l', m = k = 0 \\
0 & o.w.
\end{cases}
\end{equation}
It follows that the operator $\myvec{P}_{\text{shared}}$ is equal to,
\begin{equation}\label{eq:shared_product}
\myvec{P}_{\text{shared}} = \myvec{L}_x \myvec{L}_y +\myvec{L}_y \myvec{L}_x = \sum_l (\myvec{\phi}^s_l \otimes \myvec{\phi}^a_0) (\myvec{\phi}^s_l \otimes \myvec{\phi}^b_0)^T,
\end{equation}
where $\otimes$ denotes the Hadamard product. 
The vectors $\myvec{\phi}^a_0,\myvec{\phi}^b_0$ constitute the degree of the different observations and have little effect on the outcome. Thus, the leading eigenvectors of $\myvec{P}_{\text{shared}}$ are associated with the shared component and not the differential components in the product of manifolds. Below, we illustrate this phenomenon with two examples.



\paragraph{Example 1: points in a 3D cube.}
Consider points generated uniformly at random over a 3D cube of dimensions $[0,l_s] \times [0,l_a] \times [0,l_b]$. Let $\myY \in \R^{n \times 2}$ constitute the first two coordinates of $n$ independent observations, and let $\myX$ constitute the first and third coordinates. 
This is a simple case of a product of manifolds, where the shared variable $\theta_s$ is the first coordinate, while the modality-specific variables $\theta_a,\theta_b$ are the second and third coordinates.
Following Eq. \eqref{eq:rectangle}, the eigenvectors of the graph Laplacian matrices $\myvec{L}_x, \myvec{L}_y$, evaluated at  $(\theta_s,\theta_b)$ and $(\theta_s,\theta_a)$ converge to,
\begin{align}
&\phi_{lm}^x(\theta_s,\theta_b) = \cos(\pi l \theta_s/l_s)\cos(\pi m \theta_b/l_b) \notag \\
&\phi_{lk}^y(\theta_s,\theta_a) = \cos(\pi l \theta_s/l_s)\cos(\pi k \theta_a/l_a).
\end{align}
The first row of Fig. 1 (Appendix A) shows a scatter plot of the points in $\myvec{X}$ (located according to the first two coordinates), colored by the values of the leading eigenvectors of $\myvec{L}_x$. 
The second row shows the points in $\myvec{X}$, but colored by the eigenvectors of $\myvec{P}_{\text{shared}}$. As expected, all the eigenvectors of $\myvec{P}_{\text{shared}}$ are functions of the shared coordinate $\theta_s$. 

\paragraph{Example 2: videos taken from different angles.}
Our second example is based on an experiment done in \cite{lederman2014common}, where the two modalities constitute two videos of three dolls rotating at different angular speeds. The first camera (modality $\myX$) captures the middle and left doll, while the second camera (modality $\myY$) captures the middle and right dolls (see Fig. \ref{fig:video_samples}). Here, the shared variable $\myvec{\theta}_s$ is the angle of the middle doll captured by both modalities. The modality-specific variables $\myvec{\theta}_a,\myvec{\theta}_b$ are the angles of the left and right dolls captured by each modality separately.

%

To illustrate Eq. \eqref{eq:shared_product} in this example, we first compute an approximation of the eigenvectors $\myvec{\phi}_l^s$. To that end, we cropped each image in one of the videos such that only the middle doll (which appears in both modalities) is shown. One may think of this operation as a projection to the shared manifold. Next, we computed from the cropped images the leading eigenvectors $\myvec{\phi}^s_l$ of the Laplacian matrix. 
Fig. 2 (Appendix A) shows the
leading three eigenvectors of $\myvec{P}_{\text{shared}}$ as a function of 
$\myvec{\phi}^s_1,\myvec{\phi}^s_2,\myvec{\phi}^s_3$ as computed by the cropped images. The figure shows a linear dependency between the vectors, which implies that the shared operator retained only the shared component of the two modalities.

\subsection{The Differential Graph Operators}\label{sec:diff_op}

We design two operators $\myvec{Q}_{x}$ and $\myvec{Q}_{y}$ to infer latent structures that are \textit{modality specific} to $\myX,\myY$ respectively. 
\begin{equation}
    \myvec{Q}_{x} = \tilde{\myvec{L}}_y^{-1} \myvec{L}_x  \tilde{\myvec{L}}_y ^{-1},
\qquad
    \myvec{Q}_{y} = \tilde{\myvec{L}}_x^{-1} \myvec{L}_y  \tilde{\myvec{L}}_x^{-1} 
    \label{eq:diff_op_y},
\end{equation}
where $\tilde{\myvec{L}}_x = \myvec{L}_x + c\myvec{I}$, $\tilde{\myvec{L}}_y = \myvec{L}_y + c\myvec{I}$, and $c$ is a regularization constant. We address the cluster example used for the shared operator to motivate the use of these operators.

\paragraph{Differential clusters.}
In the synthetic cluster example in Fig. \ref{fig:vis_joint_op}, modality $\myX$ has three smaller clusters not observed in modality $\myY$.
We show that one can detect the \textit{differential clusters} of modality $\myX$ via the leading eigenvectors of $\myvec{Q}_x$.
By Eq. \eqref{eq:Laplacian_rewrite}, we can approximate $\tilde{\myvec{L}}_y$ via,
\begin{align}
        \tilde{\myvec{L}}_y &= (1+c)\myvec{V}_s \myvec{V}_{s}^{T}  + c\myvec{V}_{\text{comp}}\myvec{V}_{\text{comp}}^{T},  \label{eq:Laplacian_tilde_rewrite}
\end{align}

where $\myvec{V}_{\text{comp}} \in \mathbb{R}^{n\times (n-3)}$ contains, as columns, vectors that span the complementary subspace to $\myvec{V}_{s}$. We write $\myvec{Q}_x$ as:
\begin{equation}
    \myvec{Q}_x = \tilde{\myvec{L}}_y^{-1} \myvec{L}_x  \tilde{\myvec{L}}_y ^{-1}  =  
    c^{-2} \myvec{V}_x  \myvec{V}_x^T +  (1+c)^{-2}\myvec{V}_s \myvec{V}_s^{T}.   
   \label{eq:Qxy_rewrite}
\end{equation}


The differential operator in Eq. \eqref{eq:Qxy_rewrite} has two terms. The first spans the subspace corresponding to the differential structure $\myvec{V}_x$, while the second spans the subspace of the shared structure $\myvec{V}_s$. 
Since $c^{-2} > (1+c)^{-2}$, it follows that the leading eigenvectors of $\myvec{Q}_x$ span the subspace of $\myvec{V}_x$.

In theory, we can directly apply these operators to learn the structures. However, in many real-world applications, e.g., single-cell multi-omic technologies, both $\myvec{X}$ and $\myvec{Y}$ can be very noisy. In particular, abundant noisy features (e.g., genes) might dominate the data, and the top eigenvectors of $\myvec{L}_x$ and $\myvec{L}_y$ might not capture the underlying structure, which would be detrimental to the learning of $\myvec{P}_{\text{shared}}$, $\myvec{Q}_x$, and $\myvec{Q}_y$. As shown in the affinity matrices on the right of Fig. \ref{fig:vis_joint_op}, the structures are less clear when many noisy features are present. Therefore, it is necessary to have a feature selection framework that can effectively remove these noisy features in our multi-modal setting. With the aforementioned DUFS feature selection framework as the foundation, we will show in the next section how we can incorporate it into our proposed operators in the multi-modal setting.

\subsection{mmDUFS}
\label{sec:mvdufs}

In this section, we describe our  framework, termed multi-modal Differential Unsupervised Feature Selection (mmDUFS)\footnote{Codes are available at https://github.com/jcyang34/mmDUFS}. We incorporates differentiable gates \cite{lindenbaum2021differentiable}
with loss functions based on the shared and differential operators, detailed in Sec. \ref{sec:joint_op} and \ref{sec:diff_op}. 
Our goal is to compute an accurate shared graph operator ($\myvec{P_{\text{shared}}}$ in Eq. \eqref{eq:composite_op}) 
and differential graph operators ($\myvec{Q}_x$ and $\myvec{Q}_y$ in Eq. \eqref{eq:diff_op_y}) 
while simultaneously selecting the informative features.
Let $\myvec{f}_x,\myvec{f}_y$  denote a feature vector in $\myX,\myY$, respectively. 
To quantify how noisy or informative the features are with respect to the shared structure, we replace the Laplacian $\myvec{L}$ in Eq. \eqref{eq:LS} with $\myvec{P_{\text{shared}}}$, which yields the shared score
$\myvec{f}_x^T \myvec{P_{\text{shared}}} \myvec{f}_x $ and
$\myvec{f}_y^T \myvec{P_{\text{shared}}} \myvec{f}_y$.
Similarly, 
$\myvec{f}_x^T\myvec{Q}_x\myvec{f}_x$ and $\myvec{f}_y^T\myvec{Q}_y\myvec{f}_y$ quantify the smoothness of these features with respect to the differential graph operators $\myvec{Q}_x$ and $\myvec{Q}_y$.
The rationale behind these generalized Laplacian Scores is similar to the original score. For instance, let $\myvec{P_{\text{shared}}} = \sum_{i=1}^{n}\lambda_{i} \myvec{u}_i \myvec{u}_i^T$ be the eigendecomposition of $\myvec{P_{\text{shared}}}$. If $\myvec{f}_x$ varies slowly with respect to the underlying shared structure, it will have a larger component projected onto the subspace of $\myvec{P_{\text{shared}}}$, thus leads to a higher score.

To learn features with high generalized Laplacian Scores and accurate graph operators, mmDUFS learns two sets of Stochastic Gates $\myvec{z}_x$ and $\myvec{z}_y$ that filter irrelevant  features in each modality. Similar to DUFS \cite{lindenbaum2021differentiable}, these stochastic gates multiply the data matrices $\myvec{X}$ and $\myvec{Y}$ to remove nuisance features, i.e., 
$\myvec{\tilde{X}} = \myvec{X}\Delta(\myvec{z}_x)$
and 
$\myvec{\tilde{Y}} = \myvec{Y} \Delta(\myvec{z}_y)$.
At each iteration, the updated graph operators ($\myvec{\tilde{P}_{\text{shared}}}$, $\myvec{\tilde{Q}}_{x}$, $\myvec{\tilde{Q}}_{y}$)  are recomputed based on the gated inputs.

mmDUFS has two modes: (i) detecting shared structures  using the shared graph operator $\myvec{\tilde{P}_{\text{shared}}}$, and (ii) detecting  modality-specific structures using the differential graph operators $\myvec{\tilde{Q}}_{x}$, and $\myvec{\tilde{Q}}_{y}$. 
To learn the shared structure and the corresponding features, we propose to optimize $\myvec{z}_x$ and $\myvec{z}_y$ by minimizing the following loss function:
\begin{align*}
    \mathcal{L}_{\text{shared}} &= -\frac{1}{n} \text{Tr}[\myvec{\tilde{X}}^T \myvec{\tilde{P}_{\text{shared}}} \myvec{\tilde{X}}] - \frac{1}{n}
    \text{Tr}[\myvec{\tilde{Y}}^T \myvec{\tilde{P}_{\text{shared}}} \myvec{\tilde{Y}}]
     \\
    &+ \lambda_x \|\myvec{z}_x \|_0 +\lambda_y \|\myvec{z}_y \|_0,
    \label{eq:mvDUFS_loss1}
\end{align*}
where the first two terms are the Shared Laplacian Scores for each modality, and the regularizers 
    $\lambda_x \|\myvec{z}_x \|_0$ and $\lambda_y \|\myvec{z}_y \|_0$ control the number of selected features for each modality, with tunable parameters $\lambda_x,\lambda_y$ that control the level of sparsity.  In Appendix B.1, we suggest a procedure to tune these regularization parameters. 
Similarly, the loss functions $\mathcal{L}_{x},\mathcal{L}_{y}$ are designed to detect features associated with structures that appear only in modality $\myX,\myY$, respectively.
\begin{align}
    \mathcal{L}_{x} &= -\frac{1}{n} \text{Tr}[\myvec{\tilde{X}}^T \myvec{Q}_{\tilde{x}} \myvec{\tilde{X}}]  
    +\lambda_x \|\myvec{z}_x \|_0 ,
\notag \\
    \mathcal{L}_{y} &= -\frac{1}{n}  \text{Tr}[\myvec{\tilde{Y}}^T \myvec{Q}_{\tilde{y}} \myvec{\tilde{Y}}]  
    + \lambda_y \|\myvec{z}_y \|_0,
\end{align}
where the first term in each loss is termed Differential Laplacian Scores. In the following section we show the usefulness of these score functions for detecting relevant features. 

\section{Results}
\label{sec:simulation}

We benchmark mmDUFS using  synthetic and real multi-modal datasets. 
For discovering the shared structures and associated features, we compare mmDUFS with the shared operator to the following variants of kernel fusion-based methods previously proposed for dimensionality reduction: (1) Matrix Concatenation (MC), where the Laplacian is computed based on a concatenated matrix of the two modalities. (2) 
Multi-modal Kernel Sum (mmKS) \cite{zhou2007spectral}, where the Laplacian is equal to $\myvec{L}_x + \myvec{L}_y$. (3) Multi-modal Kernel Product (mmKP) \cite{lindenbaum2015learning,lindenbaum2020multi}. 
where the Laplacian is equal to $\myvec{L}_x\myvec{L}_y$.


For each baseline, the $k$ features with the highest Laplacian Scores are  selected. For the synthetic datasets, we set $k$ to be the correct number of informative features. We evaluate the performance of different methods by the F1-score $\text{F1} = \text{TP}/(\text{TP}+ \frac{1}{2(\text{FP}+\text{FN})})$, where TP is the number of informative features selected by each method, FP is the number of uninformative selected features, and FN is the number of missed informative features. For the rescaled MNIST and rotating doll examples, the informative features are set to the $25\%$ pixels with the highest standard deviation.


\subsection{Synthetic Examples}

\begin{figure*}[htb!]%
    \centering
   \hspace{-4em}
    \parbox{\figrasterwd}{
    \parbox{.3\figrasterwd}{%
     \centering
      \subcaptionbox{\label{fig:mnist_samples}}{\includegraphics[height=0.14\textwidth]{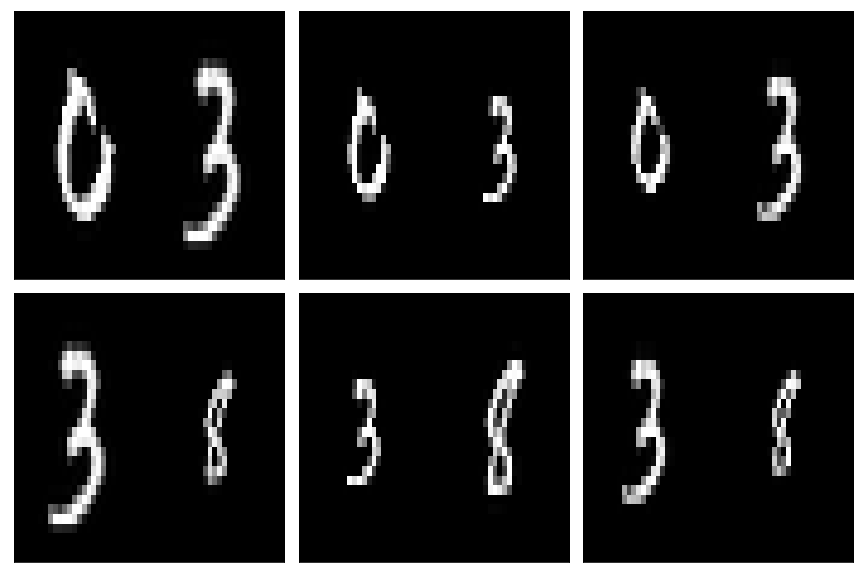}}
      \centering
      \subcaptionbox{\label{fig:mnist_gates}}{\includegraphics[height=0.14\textwidth]{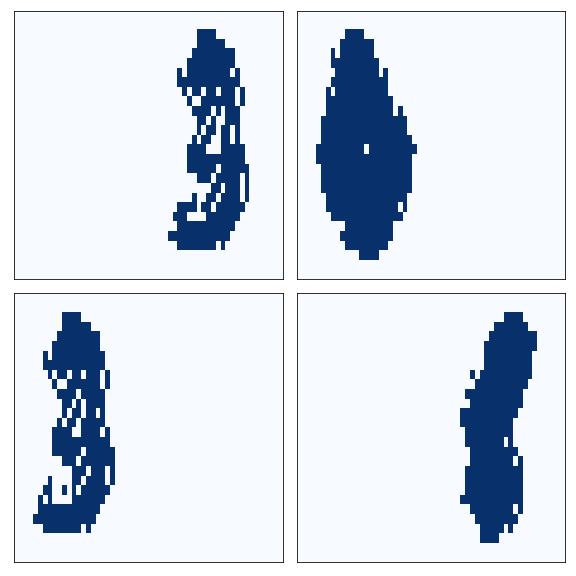}}  
    }
    \hspace{-3em}
    \parbox{.3\figrasterwd}{%
     \centering
      \subcaptionbox{\label{fig:tree_umap}}{\includegraphics[height=0.3\textwidth]{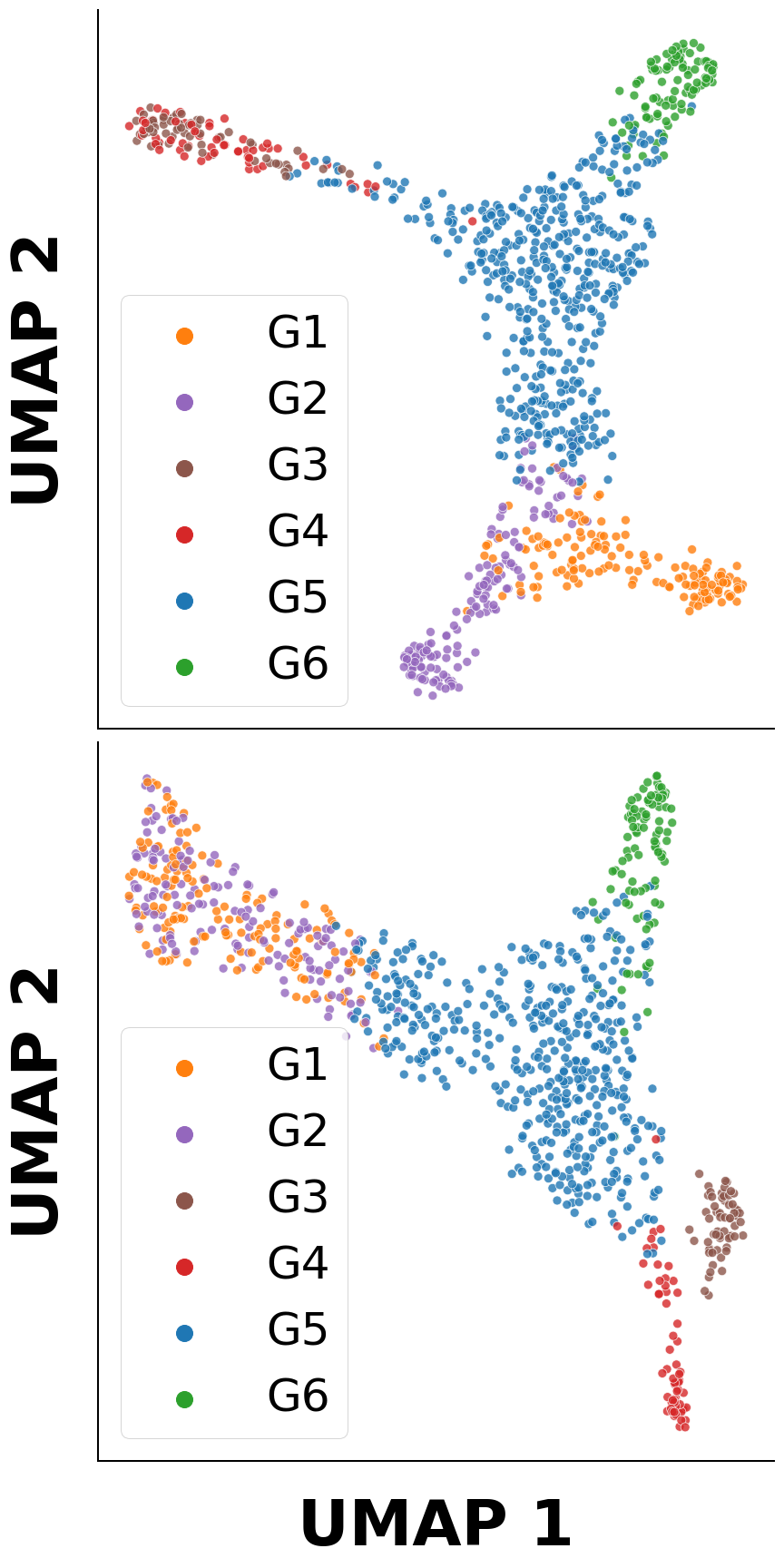}}
    }
     \hspace{-2em}
     \parbox{.3\figrasterwd}{%
      \subcaptionbox{\label{fig:tree_shared}}{\includegraphics[height=0.14\textwidth]{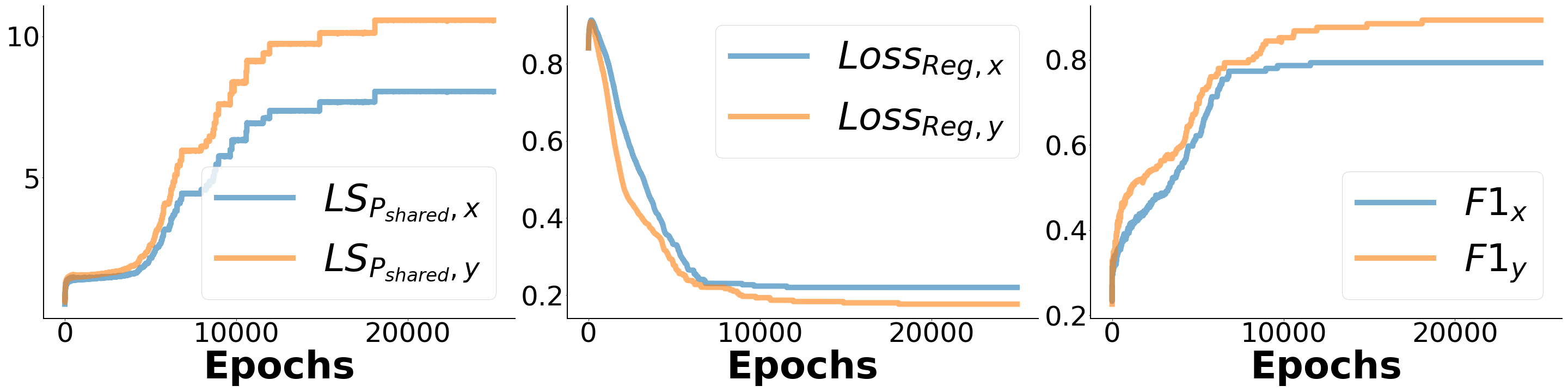}}
      \subcaptionbox{\label{fig:tree_diff}}{\includegraphics[height=0.14\textwidth]{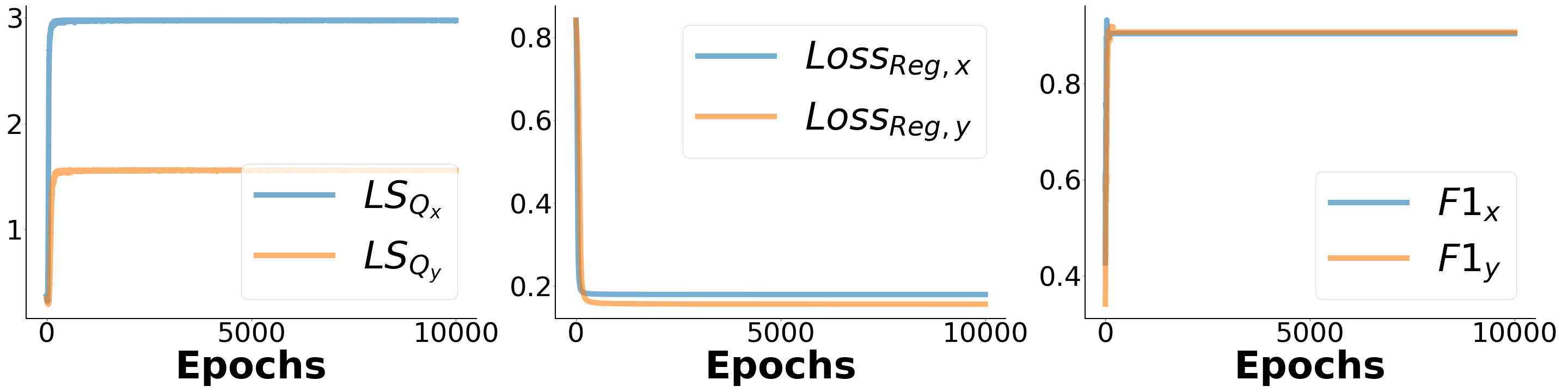}}  
    }
    }
    \caption{Left (a-b): Evaluation of the proposed approach on the rescaled MNIST dataset. (a): Random images from modality $X$ (upper row) and modality $Y$ (bottom row) in gray-scale. (b): Selected pixels (dark blue) for the shared operator (left column) and the differential operator (right column). Right (c-e): Synthetic developmental tree example. (c):  UMAP embeddings of the tree using data from modality $\myX$ (top) and modality $\myY$ (bottom). (d-e): Change of the Shared/Differential Laplacian Scores, regularization loss, and the F1-score of the selected features concerning the number of epochs (x-axis) for mmDUFS with the shared operator (panel (c)) and the differential operator (panel (e)).}%
    \label{fig:MNIST_result}%
 
\end{figure*}

\paragraph{Rescaled MNIST.} 
We designed a rescaled MNIST example with shared and modality-specific digits.
We first randomly sample one image ($28 \times 28$ pixels) of digits $0$, $3$, $8$. Then, we rescale each digit randomly and independently $500$ times resulting with $500$ images of $0$, $3$, and $8$. We concatenate pairs of $0$ and $3$ to create modality $\myX$, and pairs of the same $3$ and random $8$ to create $\myY$, see example in Fig. \ref{fig:mnist_samples}. Thus, this dataset consists of $500$ samples and $28 \times 56$ pixels in each modality, with digit $3$ shared between the modalities and digit $0$ and $8$ modality specific.

We apply mmDUFS with the shared operator to this example to select pixels corresponding to $3$. The left column of Fig. \ref{fig:mnist_gates} shows the pixels gate values from mmDUFS for  modality $\myX$ (top) and $\myY$ (bottom). We can see that selected pixels outline the shape of the digit $3$ well. Table \ref{tab:f1_all} compares the F1-score achieved by mmDUFS to three baselines. 
We can see that mmDUFS achieves a higher F1-score than all the baselines on both modalities, demonstrating its ability to identify informative features accurately.

Lastly, we apply mmDUFS with the differential operator to select modality-specific pixels. The right column of Fig. \ref{fig:mnist_gates} shows the pixel gate values for both modality $\myX$ (top) and $\myY$ (bottom). We can see that mmDUFS selects pixels that outline digits $0,8$ for modalities $\myX,\myY$, respectively. Additionally, mmDUFS achieves F1-score $0.8059$ and $0.8832$ for  $\myX$ and $\myY$, showcasing its effectiveness in identifying features contributing to the differential structures.

\begin{table}[htb!]%
       \centering
       \begin{adjustbox}{max width=0.6 \textwidth,min height = 0.7 in, valign=c}
             \begin{tabular}{|c|c||c|c|c|c|}
                \hline 
              Dataset & Modality & MC & mmKS & mmKP & mmDUFS \\
              \hline

             \multirow{2}{*}{Rescaled MNIST} &  X  &  0.3547 & 0.5291 & 0.5291 & \textbf{0.7093}\\
            
                & Y & 0.4826 & 0.6219 & 0.6219 & 
                 \textbf{0.8159}\\

            \hline
           \multirow{2}{*}{Synthetic Developmental Tree} & X & 0.6000 & 0.7800 &  0.8400 & \textbf{0.8800}\\
             &  Y &  0.7800 & 0.8000 & 0.8200 & \textbf{0.9000} \\
            \hline
              \multirow{2}{*}{Original Gaussian} & X & 0.5000 & 0.7333 & \textbf{1} & \textbf{1}\\
            & Y &  0.5500 & 0.6500 & 0.9500 & \textbf{1} \\
             
             \multirow{2}{*}{Gaussian + $10$ Noisy Feats} & X & 0.5000 & 0.7333 & \textbf{1} & \textbf{1} \\
                                               & Y &  0.5000 & 0.6500 & 0.9000 & \textbf{1} \\
            
            \multirow{2}{*}{Gaussian + $30$ Noisy Feats} & X & 0.4667 & 0.7000 & 0.9667 & \textbf{1} \\
                                              & Y & 0.4500 & 0.5500 & 0.8500 & \textbf{1} \\
           
            \multirow{2}{*}{Gaussian + $50$ Noisy Feats} & X & 0.4000 & 0.6333 & 0.9333 & \textbf{0.9667} \\
                                              & Y & 0.4000 & 0.5500 & 0.8000 & \textbf{0.8500} \\
            \hline

         \end{tabular}%
         \end{adjustbox}
    \caption{Comparison of F1-score between different methods on the rescaled MNIST example, the synthetic tree example, and the Gaussian mixture example with different numbers of additive noisy features.}%
    \label{tab:f1_all}%
 
\end{table}

\paragraph{Synthetic Developmental Tree.}

Tree structures are ubiquitous throughout different biological processes and data modalities in single-cell biology \cite{plass2018cell,zhang2021single}. To understand the interplay of different mechanisms underlying the complex developmental process, it is vital to discover the genetic features that contribute to the tree structure shared across modalities and those that contribute to modality-specific structures. 

We evaluate mmDUFS using a simulated developmental tree example generated via a tree simulator \footnote{https://github.com/dynverse/dyntoy}. The original data has $1000$ samples and $100$ features. We divide the data into half, such that each modality has $50$ informative features that contribute to the shared tree structure, as shown in the UMAP embeddings in Fig. \ref{fig:tree_umap}, where the samples in the tree are grouped into different branch groups (labeled $G_1$ to $G_6$). We then add $50$ features drawn from negative binomial distributions to each modality 
to create differential branches, that are only observed in one modality. Specifically, branches $G_1$ and $G_2$ are bifurcated in modality $\myX$ (top UMAP embeddings) but are mixed in modality $\myY$ (bottom UMAP embeddings), and $G_3$ and $G_4$ are bifurcated in modality $\myY$ but are mixed in modality $\myX$ (see Supplementary section B.3 for further details). After log transformation and z-scoring the data, we concatenate $200$ features drawn from $N(0,1)$ to each modality as noisy features.

We apply our model with the shared and differential operators to recover the features that contribute to the overall tree structure and the set of features that contribute to the split branches, respectively. Fig. \ref{fig:tree_shared} shows the change, during training with the shared loss, in the Shared/Differential Laplacian Scores, the regularization loss, and the F1-score. 
Fig. \ref{fig:tree_diff} shows the same properties for the differential loss.
Table \ref{tab:f1_all} compares the F1-score of the selected features between different methods. Here as well,  mmDUFS clearly outperforms the other methods.

\begin{figure*}[htb!]%

\hspace{-1.2em} 
   \parbox{\figrasterwd}{
    \parbox{.5\figrasterwd}{%
     \centering
      \subcaptionbox{\label{fig:video_samples}}{\includegraphics[height=0.27\textwidth]{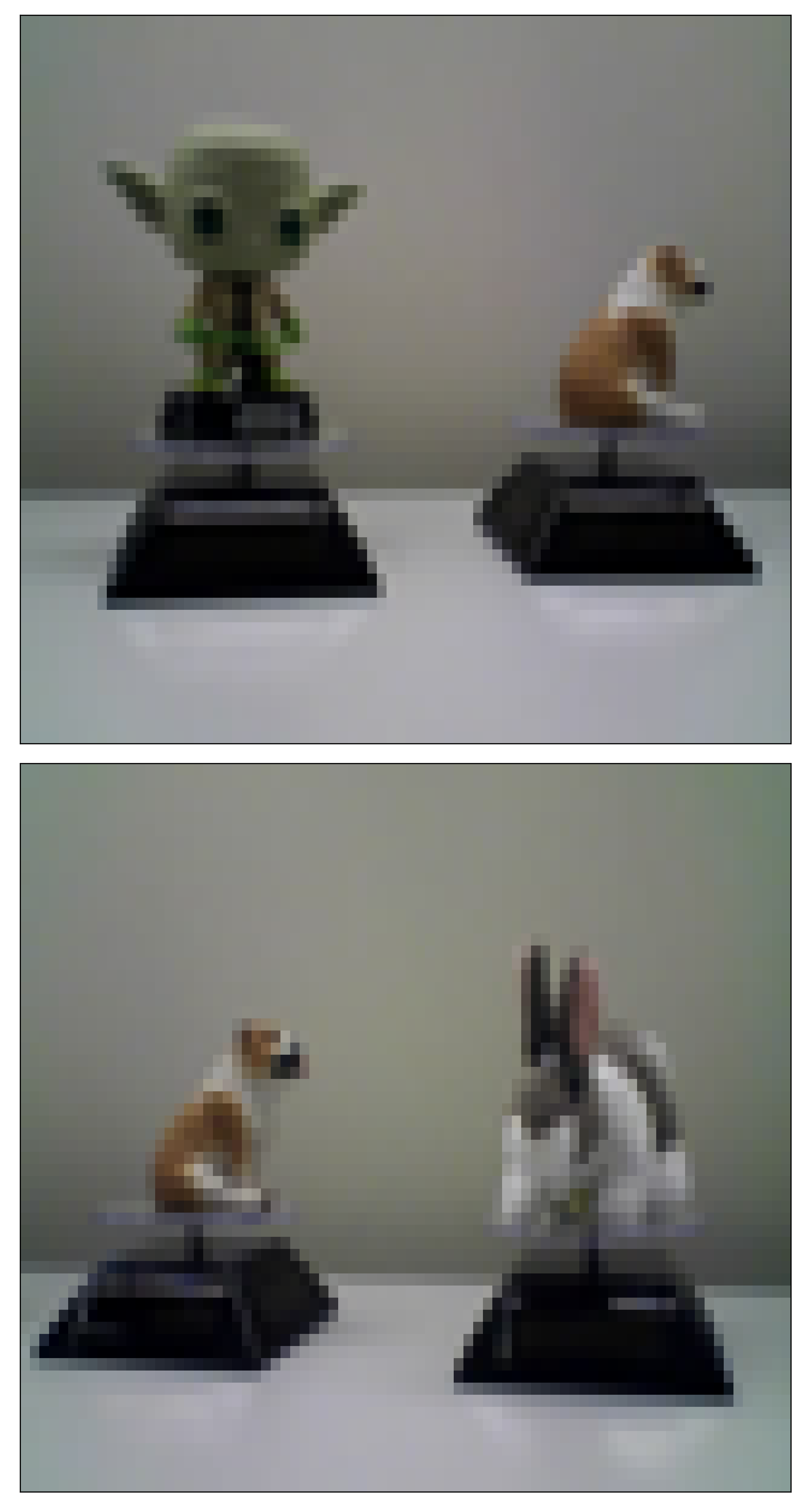}}
      \centering
      \subcaptionbox{\label{fig:video_shared}}{\includegraphics[height=0.27\textwidth]{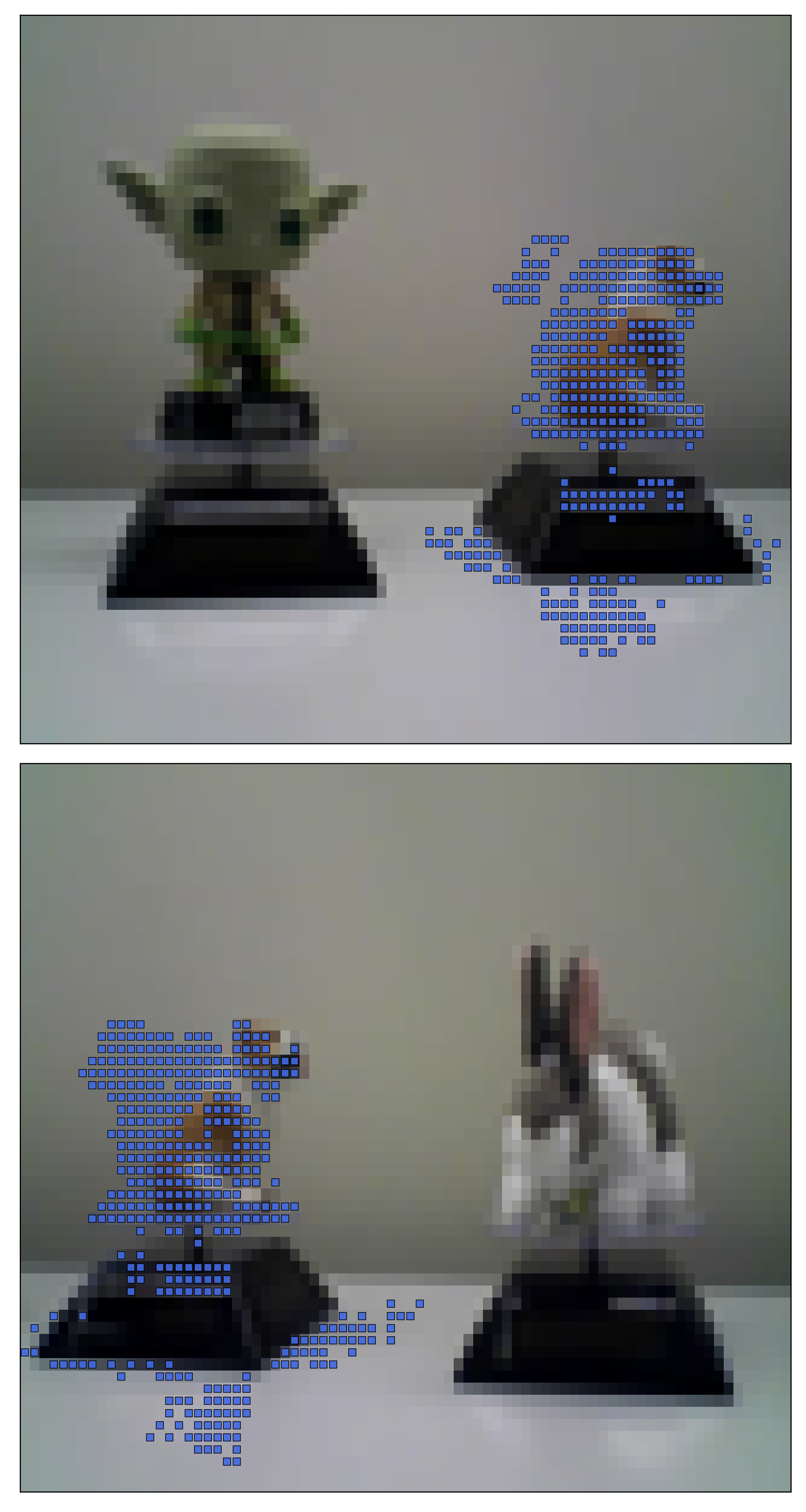}}  
     \centering
      \subcaptionbox{\label{fig:video_diff}}{\includegraphics[height=0.27\textwidth]{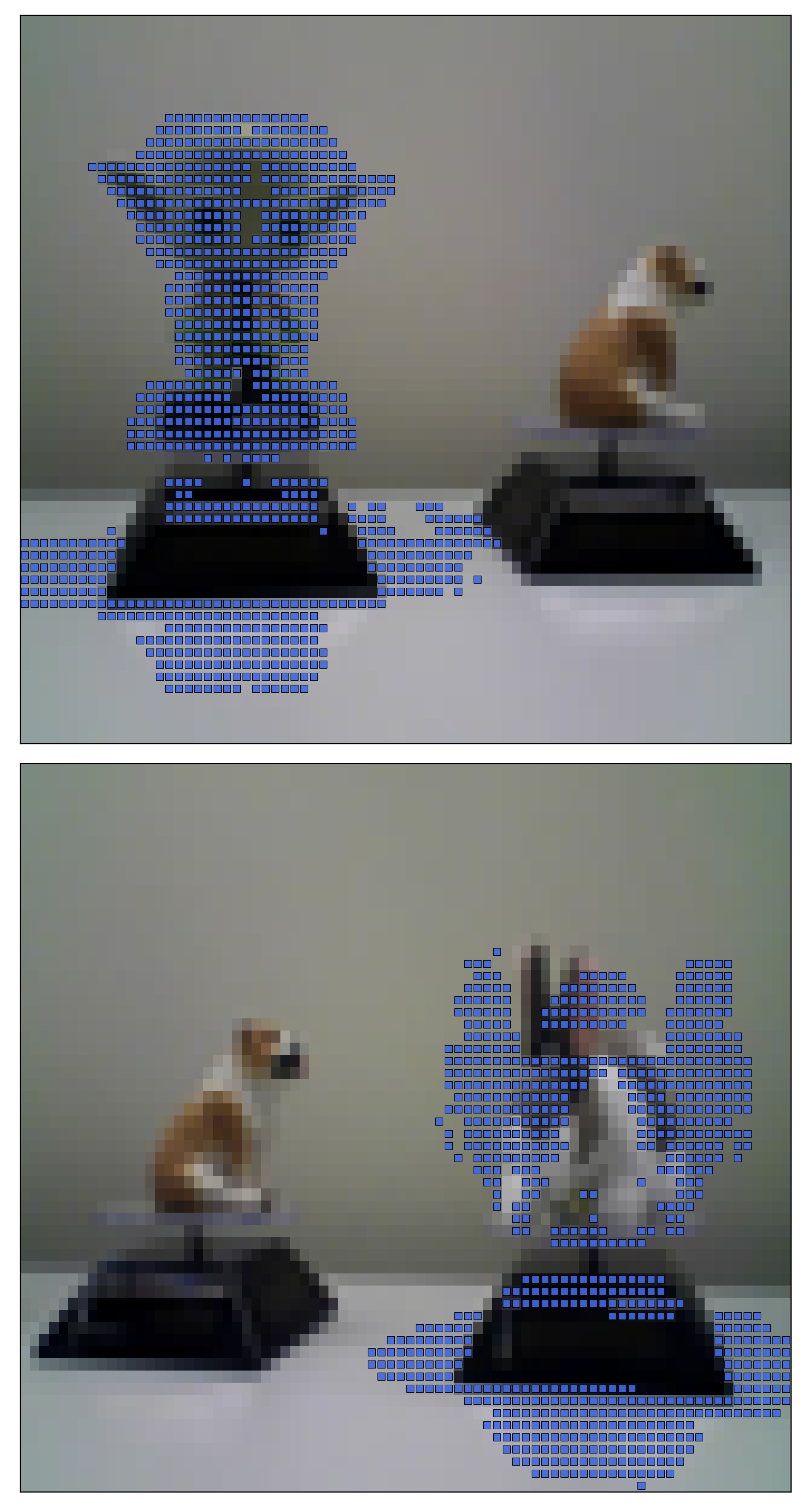}}  
      
    }
    \hspace{-4em}
    \parbox{.3\figrasterwd}{%
     \centering
      \subcaptionbox{\label{fig:cbmc_umap}}{\includegraphics[height=0.27\textwidth]{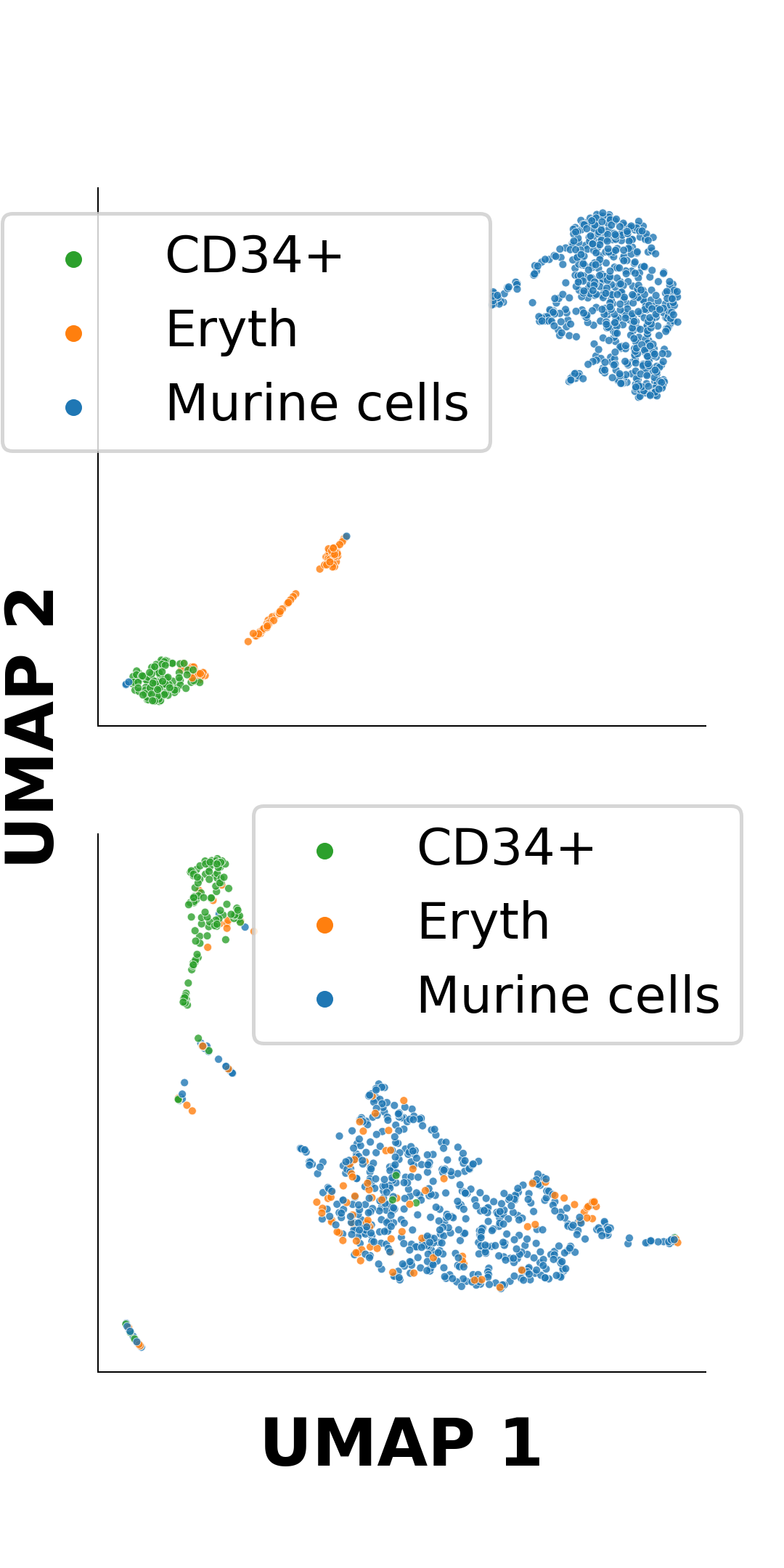}}
     
    }
     \hspace{-3.5em}
    \parbox{.3\figrasterwd}{%
      \centering
      \subcaptionbox{\label{fig:cbmc_genes}}{\includegraphics[height=0.27\textwidth]{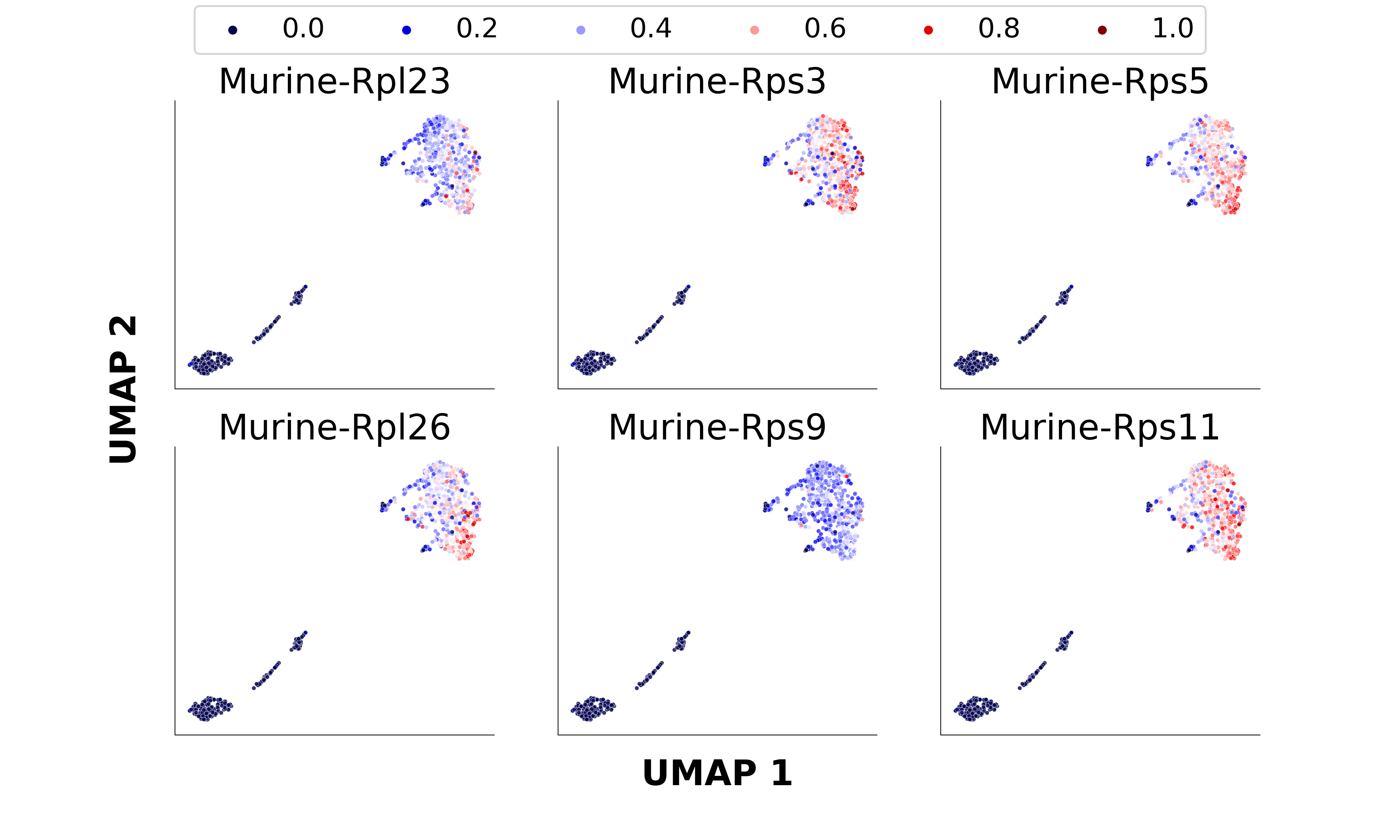}}
    }
    }

    \caption{Left (a-c): Rotating dolls example. (a): Random images of the dolls from each video. (b-c): Selected pixels are marked in blue for mmDUFS with shared operator (b) and the differential operator (c).  Right (d-e): CITE-seq data example. (d): UMAP embeddings using the RNA (top) and protein data (bottom), colored by cell type labels. (e): Similar UMAP embeddings colored by the expression level of several genes selected by mmDUFS with the differential operator. }%
    \label{fig:video_result}%
 
\end{figure*}

\paragraph{Synthetic Gaussian Mixtures.}

We generated a multi-modal Gaussian mixture dataset, where $\myX$ and $\myY$ each have $3$ clusters. Two clusters are shared between modalities, and cluster $3$ and $4$ are specific to $\myX$ and $\myY$, respectively. Each cluster has a set of informative features drawn from a multivariate Gaussian, along with noisy features (see Appendix B.2 for details). 

We first apply mmDUFS to uncover the informative features of the shared clusters and the modality-specific clusters. In the figure of Supplementary section B.2, we plot the change of the average shared/differential Laplacian Scores across features, the regularization loss, and the F1-score of the selected features from mmDUFS with respect to the number of epochs, where we can see that mmDUFS gradually selects the correct features corresponding to high scores while sparsifying the number of features. To evaluate mmDUFS's feature selection capability in challenging regimes, we further inject $10$, $30$, and $50$ noisy features into each modality and compare the F1-score of the selected features from different methods in each regime. As shown in Table \ref{tab:f1_all}, mmDUFS consistently outperforms the baseline methods while maintaining accurate feature identification capability, demonstrating its robustness against noise.

\subsection{Real Data}
\paragraph{Rotating Dolls.}



We evaluate mmDUFS's performance on the rotating doll video dataset described in Sec. \ref{sec:joint_op} in which $2$ cameras capture $2$ dolls from different angles (Fig. \ref{fig:video_samples}). By treating each video frame as one sample ($4050$ in total) and the gray-scaled pixels as features, we aim to uncover pixels that correspond to the shared doll (the dog) and the modality-specific dolls (Yoda and rabbit).

For mmDUFS with the shared operator, Fig. \ref{fig:video_shared} shows selected pixels in both videos, as indicated by the blue dots. The shape of the dog is clearly delineated in both modalities. We further compute the F1-score of the selected pixels with respect to the underlying pixels that correspond to the dog. mmDUFS achieves F1-score of $0.7158$ and $0.8033$ for the two modalities, whereas MC achieves $0.2390$ and $0.3822$, and mmKS and mmKP achieve $0.5452$ and $0.6868$. 
Fig. \ref{fig:video_diff} shows the selected pixels of mmDUFS with the differential operator in the two videos. In videos 1, mmDUFS select mostly pixels corresponding to the Yoda (F1-score: $0.8861$). For video 2, mmDUFS select mostly pixels corresponding to the rabbit (F1-score: $0.7446$).

\paragraph{CITE-seq Dataset.}
In single-cell biology, cell states are characterized by different features at different molecular levels. Identifying the contributing features is an open question crucial to understanding the underlying cell systems. We apply mmDUFS to a CITE-seq dataset from \cite{stoeckius2017simultaneous}, in which cells are profiled at both transcriptomic and proteomic levels measuring expressions of genes and protein markers, to identify the genes and proteins that characterize the cell states in the multi-modal setting.  
 
In this data, a group of murine cells is spiked-in as controls to human cord blood mononuclear cells (CBMCs), and CITE-seq sequences the resulting cell system. Fig. \ref{fig:cbmc_umap} shows  UMAP embeddings of the cells based on their RNA expression (top) and protein expression (bottom).  From the full dataset, we analyzed $3$ cell populations: murine cells (blue) and $2$ CBMCs cell populations (Erythroids (orange) and CD34+ cells (green)). This dataset has $832$ cells, with $500$ top variable genes from modality 1 and $10$ protein markers from modality 2. We can see that the murine cells are separable from the Erythroids in the RNA space but not in the proteomic space.
To identify which gene markers contribute to the separation between cell groups, we apply mmDUFS with the differential operator to this data. We found that all the selected genes are murine genes that only express in the murine cells, as shown in  Fig. \ref{fig:cbmc_genes}. This example demonstrates that mmDUFS can identify genetic markers contributing to the differential structures observed in single-cell multi-omic data.

     
 

\section{Discussion}


We present mmDUFS, a feature selection method that learns two novel graph operators that capture the \textit{shared} and the \textit{modality-specific} structures in multi-modal data, while simultaneously selecting the features that are informative for these structures. MmDUFS can operate on small batches which makes it scalable to large datasets. On the other hand, finding the optimal regularization parameters for mmDUFS on real data may be challenging, for which we suggest an automatic procedure in Appendix B.1. 
A second potential limitation is the $\mathcal O(n^3)$ computational complexity required to compute $ \tilde{\myvec{L}}$ (Eq. \eqref{eq:diff_op_y}). A possible solution is to reduce the complexity by computing a sparse Laplacian matrix. 

\section*{Acknowledgements}

The authors thank Amit Moscovich for the helpful discussions and feedback.

\bibliography{ref}

\begin{thebibliography}{10}

\bibitem{ma2020chromatin}
Sai Ma, Bing Zhang, Lindsay~M LaFave, Andrew~S Earl, Zachary Chiang, Yan Hu,
  Jiarui Ding, Alison Brack, Vinay~K Kartha, Tristan Tay, et~al.
\newblock Chromatin potential identified by shared single-cell profiling of rna
  and chromatin.
\newblock {\em Cell}, 183(4):1103--1116, 2020.

\bibitem{liu2020high}
Yang Liu, Mingyu Yang, Yanxiang Deng, Graham Su, Archibald Enninful, Cindy~C
  Guo, Toma Tebaldi, Di~Zhang, Dongjoo Kim, Zhiliang Bai, et~al.
\newblock High-spatial-resolution multi-omics sequencing via deterministic
  barcoding in tissue.
\newblock {\em Cell}, 183(6):1665--1681, 2020.

\bibitem{stoeckius2017simultaneous}
Marlon Stoeckius, Christoph Hafemeister, William Stephenson, Brian
  Houck-Loomis, Pratip~K Chattopadhyay, Harold Swerdlow, Rahul Satija, and
  Peter Smibert.
\newblock Simultaneous epitope and transcriptome measurement in single cells.
\newblock {\em Nature methods}, 14(9):865--868, 2017.

\bibitem{joung2023transcription}
Julia Joung, Sai Ma, Tristan Tay, Kathryn~R Geiger-Schuller, Paul~C
  Kirchgatterer, Vanessa~K Verdine, Baolin Guo, Mario~A Arias-Garcia, William~E
  Allen, Ankita Singh, et~al.
\newblock A transcription factor atlas of directed differentiation.
\newblock {\em Cell}, 186(1):209--229, 2023.

\bibitem{xiao2022spatially}
Yang Xiao, Graham Su, Yang Liu, Cheick~A Sissoko, Yung-yu Huang, Adrienne~N
  Santiago, Andrew~J Dwork, Gorazd~B Rosoklija, Underwood~D Mark, Victoria
  Arango, et~al.
\newblock Spatially resolved transcriptomes in human hippocampus.
\newblock {\em Biological Psychiatry}, 91(9):S18, 2022.

\bibitem{leblay2020cite}
Noemie Leblay, Ranjan Maity, Elie Barakat, Sylvia McCulloch, Peter Duggan,
  Victor Jimenez-Zepeda, Nizar~J Bahlis, and Paola Neri.
\newblock Cite-seq profiling of t cells in multiple myeloma patients undergoing
  bcma targeting car-t or bites immunotherapy.
\newblock {\em Blood}, 136:11--12, 2020.

\bibitem{sun2013survey}
Shiliang Sun.
\newblock A survey of multi-view machine learning.
\newblock {\em Neural computing and applications}, 23:2031--2038, 2013.

\bibitem{yan2021deep}
Xiaoqiang Yan, Shizhe Hu, Yiqiao Mao, Yangdong Ye, and Hui Yu.
\newblock Deep multi-view learning methods: A review.
\newblock {\em Neurocomputing}, 448:106--129, 2021.

\bibitem{raij2000audiovisual}
Tommi Raij, Kimmo Uutela, and Riitta Hari.
\newblock Audiovisual integration of letters in the human brain.
\newblock {\em Neuron}, 28(2):617--625, 2000.

\bibitem{radford2021learning}
Alec Radford, Jong~Wook Kim, Chris Hallacy, Aditya Ramesh, Gabriel Goh,
  Sandhini Agarwal, Girish Sastry, Amanda Askell, Pamela Mishkin, Jack Clark,
  et~al.
\newblock Learning transferable visual models from natural language
  supervision.
\newblock In {\em International conference on machine learning}, pages
  8748--8763. PMLR, 2021.

\bibitem{guzhov2022audioclip}
Andrey Guzhov, Federico Raue, J{\"o}rn Hees, and Andreas Dengel.
\newblock Audioclip: Extending clip to image, text and audio.
\newblock In {\em ICASSP 2022-2022 IEEE International Conference on Acoustics,
  Speech and Signal Processing (ICASSP)}, pages 976--980. IEEE, 2022.

\bibitem{CCA1}
Harold Hotelling.
\newblock Relations between two sets of variates.
\newblock {\em Biometrika}, 28(3/4):321--377, 1936.

\bibitem{DCCA}
Galen Andrew, Raman Arora, Jeff Bilmes, and Karen Livescu.
\newblock Deep canonical correlation analysis.
\newblock In {\em International Conference on Machine Learning}, pages
  1247--1255, 2013.

\bibitem{lindenbaum2022lsparse}
Ofir Lindenbaum, Moshe Salhov, Amir Averbuch, and Yuval Kluger.
\newblock L0-sparse canonical correlation analysis.
\newblock In {\em International Conference on Learning Representations}, 2022.

\bibitem{cca_bio}
Harold Pimentel, Zhiyue Hu, and Haiyan Huang.
\newblock Biclustering by sparse canonical correlation analysis.
\newblock {\em Quantitative Biology}, 6(1):56--67, 2018.

\bibitem{cca_fault}
Zhiwen Chen, Steven~X Ding, Tao Peng, Chunhua Yang, and Weihua Gui.
\newblock Fault detection for non-gaussian processes using generalized
  canonical correlation analysis and randomized algorithms.
\newblock {\em IEEE Transactions on Industrial Electronics}, 65(2):1559--1567,
  2017.

\bibitem{solorio2020review}
Sa{\'u}l Solorio-Fern{\'a}ndez, J~Ariel Carrasco-Ochoa, and Jos{\'e}~Fco
  Mart{\'\i}nez-Trinidad.
\newblock A review of unsupervised feature selection methods.
\newblock {\em Artificial Intelligence Review}, 53(2):907--948, 2020.

\bibitem{degeest2018smoothness}
Alexandra Degeest, Michel Verleysen, and Beno{\^\i}t Fr{\'e}nay.
\newblock Smoothness bias in relevance estimators for feature selection in
  regression.
\newblock In {\em Artificial Intelligence Applications and Innovations: 14th
  IFIP WG 12.5 International Conference, AIAI 2018, Rhodes, Greece, May 25--27,
  2018, Proceedings 14}, pages 285--294. Springer, 2018.

\bibitem{he2005laplacian}
Xiaofei He, Deng Cai, and Partha Niyogi.
\newblock Laplacian score for feature selection.
\newblock {\em Advances in neural information processing systems}, 18, 2005.

\bibitem{zhao2012spectral}
Zheng~Alan Zhao and Huan Liu.
\newblock {\em Spectral feature selection for data mining}.
\newblock Taylor \& Francis, 2012.

\bibitem{shaham2021deep}
Uri Shaham, Ofir Lindenbaum, Jonathan Svirsky, and Yuval Kluger.
\newblock Deep unsupervised feature selection by discarding nuisance and
  correlated features.
\newblock {\em Neural Networks}, 152:34--43, 2022.

\bibitem{balin2019concrete}
Muhammed~Fatih Bal{\i}n, Abubakar Abid, and James Zou.
\newblock Concrete autoencoders: Differentiable feature selection and
  reconstruction.
\newblock In {\em International conference on machine learning}, pages
  444--453. PMLR, 2019.

\bibitem{yamada2020feature}
Yutaro Yamada, Ofir Lindenbaum, Sahand Negahban, and Yuval Kluger.
\newblock Feature selection using stochastic gates.
\newblock In {\em International Conference on Machine Learning}, pages
  10648--10659. PMLR, 2020.

\bibitem{yang2022locally}
Junchen Yang, Ofir Lindenbaum, and Yuval Kluger.
\newblock Locally sparse neural networks for tabular biomedical data.
\newblock In {\em International Conference on Machine Learning}, pages
  25123--25153. PMLR, 2022.

\bibitem{lindenbaum2021differentiable}
Ofir Lindenbaum, Uri Shaham, Erez Peterfreund, Jonathan Svirsky, Nicolas Casey,
  and Yuval Kluger.
\newblock Differentiable unsupervised feature selection based on a gated
  laplacian.
\newblock {\em Advances in Neural Information Processing Systems}, 34, 2021.

\bibitem{linderman2019fast}
George~C Linderman, Manas Rachh, Jeremy~G Hoskins, Stefan Steinerberger, and
  Yuval Kluger.
\newblock Fast interpolation-based t-sne for improved visualization of
  single-cell rna-seq data.
\newblock {\em Nature methods}, 16(3):243--245, 2019.

\bibitem{peterfreund2020local}
Erez Peterfreund, Ofir Lindenbaum, Felix Dietrich, Tom Bertalan, Matan Gavish,
  Ioannis~G Kevrekidis, and Ronald~R Coifman.
\newblock Local conformal autoencoder for standardized data coordinates.
\newblock {\em Proceedings of the National Academy of Sciences},
  117(49):30918--30927, 2020.

\bibitem{belkin2003laplacian}
Mikhail Belkin and Partha Niyogi.
\newblock Laplacian eigenmaps for dimensionality reduction and data
  representation.
\newblock {\em Neural computation}, 15(6):1373--1396, 2003.

\bibitem{von2007tutorial}
Ulrike Von~Luxburg.
\newblock A tutorial on spectral clustering.
\newblock {\em Statistics and computing}, 17(4):395--416, 2007.

\bibitem{shnitzer2019recovering}
Tal Shnitzer, Mirela Ben-Chen, Leonidas Guibas, Ronen Talmon, and Hau-Tieng Wu.
\newblock Recovering hidden components in multimodal data with composite
  diffusion operators.
\newblock {\em SIAM Journal on Mathematics of Data Science}, 1(3):588--616,
  2019.

\bibitem{zhang2021product}
Sharon Zhang, Amit Moscovich, and Amit Singer.
\newblock Product manifold learning.
\newblock In {\em International Conference on Artificial Intelligence and
  Statistics}, pages 3241--3249. PMLR, 2021.

\bibitem{cheng2022eigen}
Xiuyuan Cheng and Nan Wu.
\newblock Eigen-convergence of gaussian kernelized graph laplacian by manifold
  heat interpolation.
\newblock {\em Applied and Computational Harmonic Analysis}, 61:132--190, 2022.

\bibitem{garcia2020error}
Nicol{\'a}s Garc{\'\i}a~Trillos, Moritz Gerlach, Matthias Hein, and Dejan
  Slep{\v{c}}ev.
\newblock Error estimates for spectral convergence of the graph laplacian on
  random geometric graphs toward the laplace--beltrami operator.
\newblock {\em Foundations of Computational Mathematics}, 20(4):827--887, 2020.

\bibitem{lederman2014common}
Roy~R Lederman and Ronen Talmon.
\newblock Common manifold learning using alternating-diffusion.
\newblock {\em submitted, Tech. Report YALEUIDCSITR1497}, 2014.

\bibitem{zhou2007spectral}
Dengyong Zhou and Christopher~JC Burges.
\newblock Spectral clustering and transductive learning with multiple views.
\newblock In {\em Proceedings of the 24th international conference on Machine
  learning}, pages 1159--1166, 2007.

\bibitem{lindenbaum2015learning}
Ofir Lindenbaum, Arie Yeredor, and Moshe Salhov.
\newblock Learning coupled embedding using multiview diffusion maps.
\newblock In {\em Latent Variable Analysis and Signal Separation: 12th
  International Conference, LVA/ICA 2015, Liberec, Czech Republic, August
  25-28, 2015, Proceedings 12}, pages 127--134. Springer, 2015.

\bibitem{lindenbaum2020multi}
Ofir Lindenbaum, Arie Yeredor, Moshe Salhov, and Amir Averbuch.
\newblock Multi-view diffusion maps.
\newblock {\em Information Fusion}, 55:127--149, 2020.

\bibitem{plass2018cell}
Mireya Plass, Jordi Solana, F~Alexander Wolf, Salah Ayoub, Aristotelis Misios,
  Petar Gla{\v{z}}ar, Benedikt Obermayer, Fabian~J Theis, Christine Kocks, and
  Nikolaus Rajewsky.
\newblock Cell type atlas and lineage tree of a whole complex animal by
  single-cell transcriptomics.
\newblock {\em Science}, 360(6391):eaaq1723, 2018.

\bibitem{zhang2021single}
Kai Zhang, James~D Hocker, Michael Miller, Xiaomeng Hou, Joshua Chiou,
  Olivier~B Poirion, Yunjiang Qiu, Yang~E Li, Kyle~J Gaulton, Allen Wang,
  et~al.
\newblock A single-cell atlas of chromatin accessibility in the human genome.
\newblock {\em Cell}, 184(24):5985--6001, 2021.

\end{thebibliography}
\bibliographystyle{unsrt}

\newpage
\appendix
\renewcommand\thefigure{\thesection.\arabic{figure}} 
\setcounter{figure}{0}  
\renewcommand\thetable{\thesection.\arabic{table}} 
\setcounter{table}{0}  

\onecolumn
\section{Additional Simulation Results}\label{sec:extra_simulations}

\subsection{Points in a 3D cube.}
\begin{figure}[htp!]
    \centering
    \includegraphics[width = 0.95\linewidth]{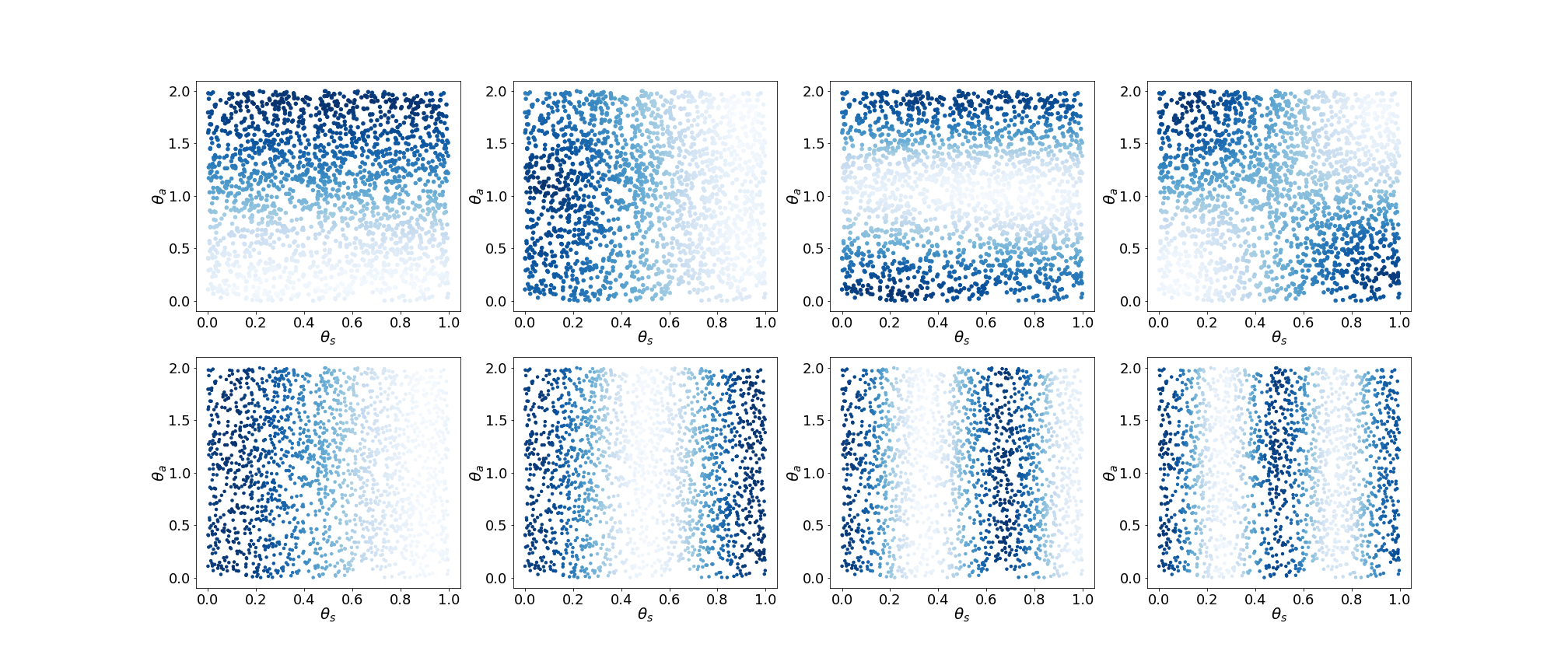}
    \caption{Data consists of points sampled uniformly at random in a 3D cube. The upper row shows a scatter plot of the points, located according to the first two coordinates $a,b$ and colored by the leading eigenvectors of $\myvec{L}_x$, the Laplacian matrix of modality $\myX$. The bottom row shows the leading eigenvectors of $\myvec{P}_{\text{shared}}$, the product of Laplacians as defined in Eq. \ref{eq:composite_op}.}
    \label{fig:3d_cube_eigenvectors}
\end{figure}

The data consists of points in a 3D cube $[0,l_s] \times [0,l_a] \times [0,l_b]$. The modality $\myX$ includes the first two coordinates, and modality $\myY$ includes the first and third, as explained in Sec. \ref{sec:method}. The upper row in Figure \ref{fig:3d_cube_eigenvectors} shows the eigenvectors of $\myvec{L}_x$. The eigenvectors change in both coordinates. The second row contains the eigenvectors of $\myvec{P}_{\text{shared}}$. the leading eigenvectors change only with the first coordinate, as it is the only shared variable.

\subsection{Rotating Dolls.}
The two modalities include video frames taken simultaneously from two cameras, of three dolls rotating at different angular speeds. The first camera (modality $\myX$) captures the left two dolls while the right camera (modality $\myY$) captures the right two dolls. Thus, the angle of the middle doll constitutes a shared variable $\myvec{\theta}_s$. The angle of the left doll $\myvec{\theta}_x$ is modality $\myX$-specific latent variable, and the angle of the right doll $\myvec{\theta}_y$ is modality $\myY$-specific latent variable.

\begin{figure}[htp!]
    \centering
    \includegraphics[width = 0.9\linewidth]{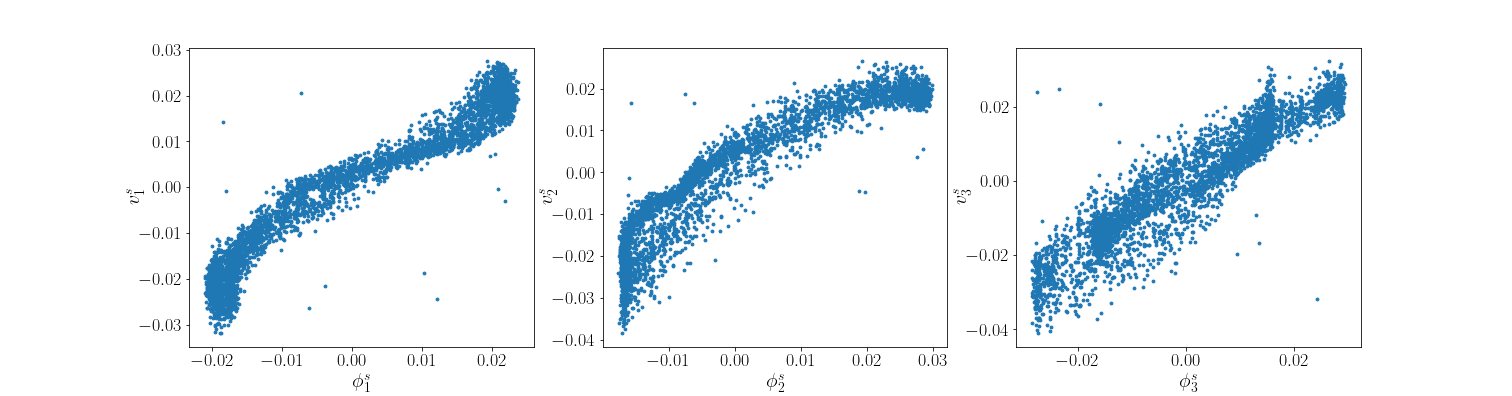}
    \caption{The figure shows a scatter plot of $\myvec{v}^s_i$, the leading eigenvectors of $\myvec{P_{\text{shared}}}$ as a function of $\myvec{\phi}^s_i$, the estimated leading vectors of the shared component in the rotating doll dataset.}
    \label{fig:rotating_dolls_operator}
\end{figure}

From the left video, we cut the frames such that it includes only the middle doll (the shared component). From these images, we computed a graph Laplacian matrix and its leading eigenvectors denoted $\myvec{\phi}^s_i$. As explained in Sec. \ref{sec:method}, we expect the eigenvectors of the shared operator, denoted $\myvec{v}_i^s$ to be similar to $\myvec{\phi}^s_i$, as both are associated with the latent variable $\myvec{\theta}_s$. Figure \ref{fig:rotating_dolls_operator} shows $\myvec{v}_i^s$ as a function of $\myvec{\phi}_i^s$
for $i=1,2,3$. The three vectors are clearly highly correlated.

\subsection{Synthetic Gaussian Mixtures.}

\begin{figure*}[htb!]%
    \centering
      \parbox{\figrasterwd}{
    \parbox{.4\figrasterwd}{%
      \subcaptionbox{\label{fig:gauss_mat}}{\includegraphics[height=0.4\textwidth]{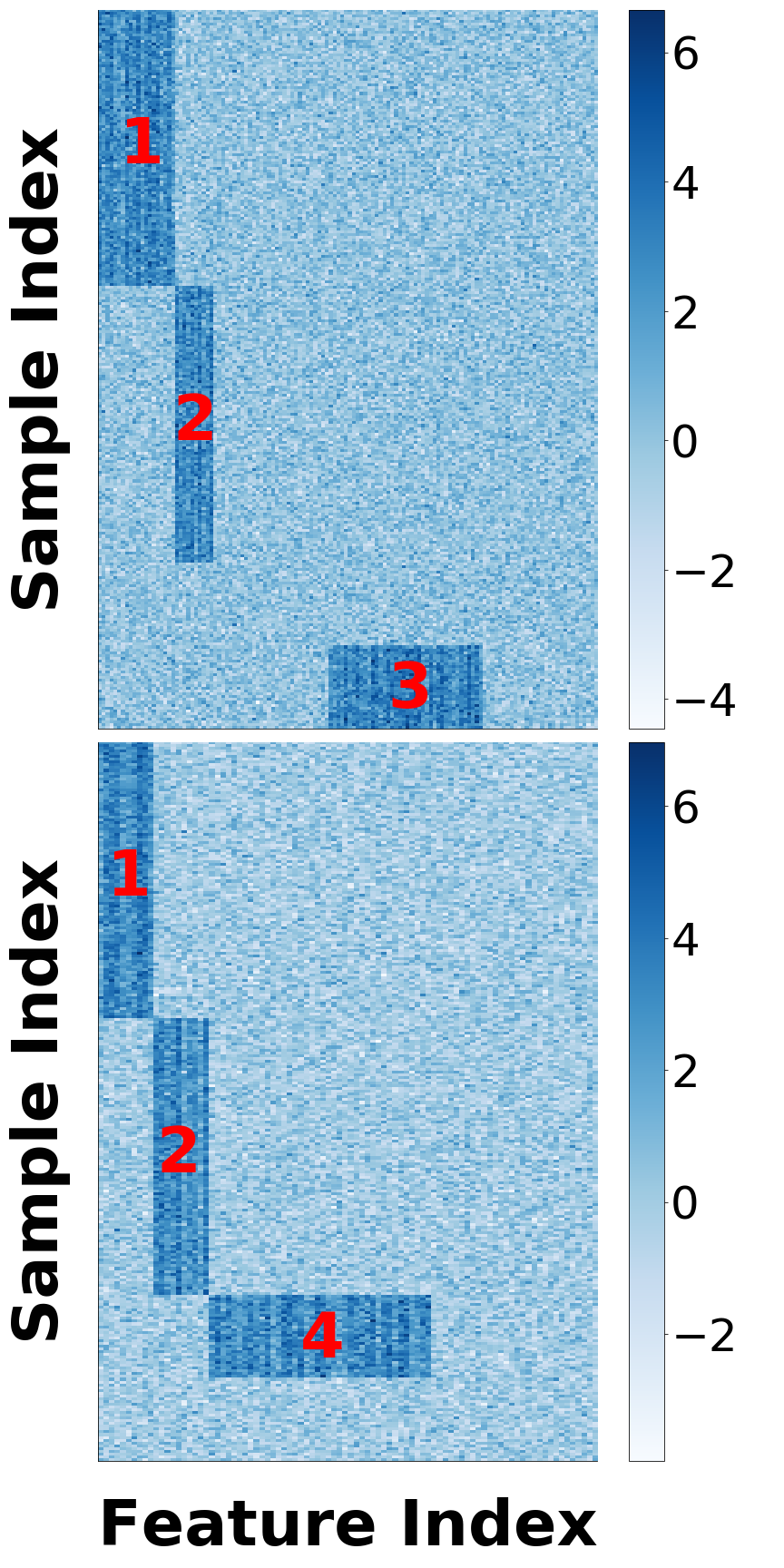}}}
    \hspace*{-7em}
    \parbox{.3\figrasterwd}{%
      \subcaptionbox{\label{fig:gauss_shared}}{\includegraphics[height=0.17\textwidth]{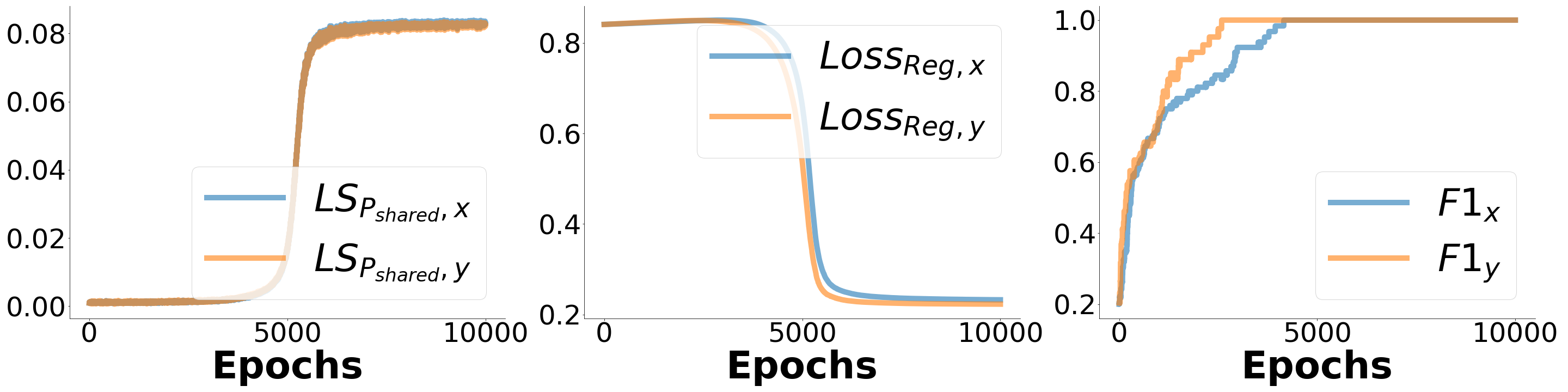}}
      \vskip1em
      \subcaptionbox{\label{fig:gauss_diff}}{\includegraphics[height=0.17\textwidth]{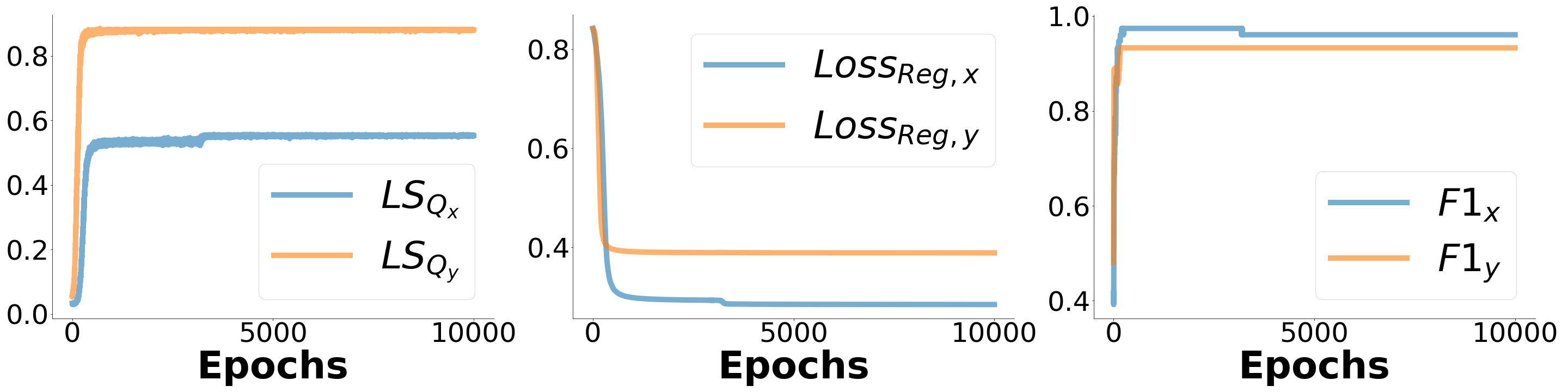}}  
    }}
    
    \vskip -0.05 in
    \caption{Synthetic Gaussian mixture cluster example. (a): Data matrix of modality $\myvec{X}$ (top) and $\myvec{Y}$ (bottom). Rows are samples, and columns are features. Each modality has $3$ clusters (labeled in red). Clusters $1$ and $2$ are shared between modalities, and cluster $3$ and $4$ are specific to each modality. (b): Change of the Shared Laplacian Scores, regularization loss, and the F1-score of the selected features concerning the number of epochs (x-axis) for mmDUFS with the shared operator. (c): Change of the Differential Laplacian Scores, regularization loss, and the F1-score of the selected features concerning the number of epochs (x-axis) for mmDUFS with the differential operator.}%
    \label{fig:supp_gauss_data}%
 
\end{figure*}

Here we apply mmDUFS to uncover the informative features of the shared clusters and the modality-specific clusters. Fig. \ref{fig:gauss_shared} and Fig. \ref{fig:gauss_diff} show the change of the average Shared/Differential Laplacian Scores across features, the regularization loss, and the F1-score of the selected features from mmDUFS with respect to the number of epochs, where we can see that mmDUFS gradually selects the correct features corresponding to high scores while sparsifying the number of features. 

\section{Experiment Details}\label{sec:exp_details}

In the following subsections, we provide additional experimental details required for the reproduction of the experiments provided in the main text. The CPU model used for the experiments is Intel(R) Xeon(R) Gold 6150 CPU @ 2.70GHz (72 cores total). The GPU model is NVIDIA GeForce RTX 2080 Ti. 

Below in Table \ref{tab:shared_param} and \ref{tab:diff_param}, we list the parameters we used on each experiment for mmDUFS with the shared operator and the differential operator. Parameter $c$ is a regularization constant for mmDUFS with the differential operator, as mentioned in the main text. Parameter $b$ is a scaling factor to the operators to balance between the Shared/Differential Laplacian Scores with respect to the regularization term. We used normalized Laplacian Matrix throughout the experiments except for the CITE-seq example where we found the performance was satisfactory with the un-normalized Laplacian Matrix.

\begin{table}[htb!]
    \centering
    \begin{tabular}{|c|c|c|c|c|c|}
      \hline
    Datasets & learning rate & epochs & $\lambda_x$  & $\lambda_y$ & $b$  \\
    \hline
    Rescaled MNIST & $2$ & $10000$ & $1e-1$ & $1e-1$ & $1e2$\\
    \hline
    Synthetic Tree & $2$ & $25000$ & $1e-1$ & $1e-1$ & $1e3$\\
    \hline
    Gaussian Mixture & $2$ & $10000$ & $1e-4$ & $1e-4$ &  $1$\\
    \hline
    Gaussian Mixture ($10$ Noisy Features) & $2$ & $20000$ & $1e-8$ & $1e-6$ &  $1$\\
    \hline
   Gaussian Mixture ($30$ Noisy Features) & $2$ & $40000$ & $1e-4$ & $1e-4$ &  $1$\\
    \hline
    Gaussian Mixture ($50$ Noisy Features) & $2$ & $10000$ & $1e-2$ & $1e-3$ &  $1e2$\\
    \hline
    Rotating Dolls & $2$ & $10000$ & $0.2$  & $0.2$ &  $1e3$\\
    \hline
    \end{tabular}
    \caption{Parameters for mmDUFS with the shared operator across different datasets.}
    \label{tab:shared_param}
\end{table}

\begin{table}[htb!]
    \centering
    \begin{tabular}{|c|c|c|c|c|c|c|}
      \hline
    Datasets & learning rate & epochs & $\lambda_x$  & $\lambda_y$ &$c$ & $b$  \\
    \hline
    Rescaled MNIST & $1$ & $10000$ & $0.5$ & $0.5$ & $1e-3$ & $1e-4$\\
    \hline
    Synthetic Tree & $2$ & $10000$ & $4$ & $2$ & $1e-3$ & $1e-3$\\
    \hline
    Gaussian Mixture & $1$ & $10000$ & $0.4$ & $0.4$ & $1e-1$ &  $1e-1$\\
    \hline
    Rotating Dolls & $2$ & $10000$ & $2$  & $2$ & $3$ &  $1e3$\\
    \hline
    CITE-seq & $2$ & $5000$ & $3$  & \diagbox & $2$ &  $1$\\
    \hline
    \end{tabular}
    \caption{Parameters for mmDUFS with the differential operator across different datasets.}
    \label{tab:diff_param}
\end{table}

For the baseline methods, $k$ features with the highest Laplacian Scores are selected. When evaluating the F1-score on the synthetic datasets, we set $k$ to be the correct number of informative features. To make a fair comparison, we also let mmDUFS select $k$ features by sorting the raw gates ($\mu_d$ for feature $d$). For other datasets, we define selected features by mmDUFS as features whose gates converged to $1$ ($z_d = 1$ for feature $d$). 

For the image datasets (rescaled MNIST, rotating dolls), we add small Gaussian noise drawn from $N(0,\sigma^2)$ to the pixels to stabilize feature selection of mmDUFS. For the rescaled MNIST dataset, $\sigma 
 = 0.1$ and we add noise to the non-informative pixels before standardizing the pixels via z-scoring. For the rotating dolls data, $\sigma = 5e-3$ and we add noise to all pixels before standardizing the pixels via z-scoring.

\subsection{Tuning of the Regularization Parameter}

mmDUFS has tunable regularization parameters $\lambda_x$ and $\lambda_y$ that control the sparsity of the number of selected features. For synthetic datasets, one can tune these parameters to select features such that the selected number is close to the prescribed number $s$. However, it can still be time and resource-consuming to optimize these parameters. Also, for real data, one might not know how many features to select and what $\lambda_x$ and $\lambda_y$ to choose.

To alleviate this issue, we propose a "warm-up" procedure similar to \cite{lindenbaum2021differentiable} to optimize $\lambda_x$ and $\lambda_y$. Specifically, we evaluate the mean Shared Laplacian Scores $S_{\text{shared}} = \frac{1}{2n} (\text{Tr}[\myvec{\tilde{X}}^T \myvec{\tilde{P}_{\text{shared}}} \myvec{\tilde{X}}]/{m} +  \text{Tr}[\myvec{\tilde{Y}}^T \myvec{\tilde{P}_{\text{shared}}} \myvec{\tilde{Y}}]/{d})$  and the mean Differential Laplacian Scores  $S_{\text{x}} = \text{Tr}[\myvec{\tilde{X}}^T \myvec{Q}_{\tilde{x}} \myvec{\tilde{X}}]/{(d\times n)}$, $S_{\text{y}} = \text{Tr}[\myvec{\tilde{Y}}^T \myvec{Q}_{\tilde{y}} \myvec{\tilde{Y}}]/{(m \times n)}$ over a grid of $\lambda_x$ and $\lambda_y$ at the early stage of training (e.g., first $1000$ epochs), and pick the parameters that maximize the Scores. Here $n$ is the number of samples in the batch, and $m$ and $d$ are the number of selected features on each modality for real data or the number of pre-specified features for synthetic data.

\begin{figure}
    \centering
    \includegraphics[width = 0.8\linewidth]{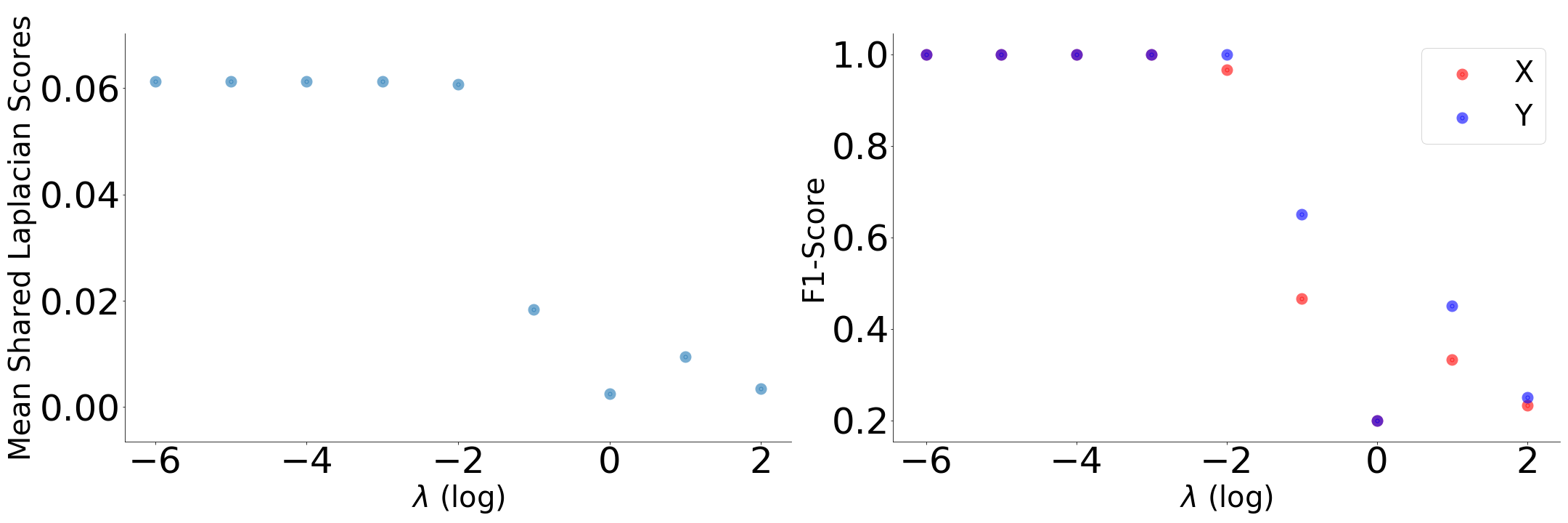}
    \caption{Evaluation of the mean Shared Laplacian Scores (left) and the corresponding F1-scores (right) over a grid of $\lambda$s on the synthetic Gaussian mixture dataset. the $y$-axis shows the mean Shared Laplacian Scores (left) and F1-scores (right) whereas the $x$-axis shows the values of $\lambda$.}
    \label{fig:gauss_lambda}
\end{figure}

To demonstrate this procedure, we use the synthetic Gaussian mixture dataset as the example, and we evaluate $\lambda_x$ and $\lambda_y$ over \{$1e-6$,$1e-5$,$1e-4$,$1e-3$,$1e-2$,$1e-1$,$1$,$1e1$,$1e2$\} using mmDUFS with the shared operator. For illustration purpose, we set $\lambda_x = \lambda_y$ Fig. \ref{fig:gauss_lambda} shows the mean Shared Laplacian Scores over different $\lambda$ values. We can see that \{$1e-6$,$1e-5$,$1e-4$,$1e-3$\} are the best candidates that give the highest Shared Laplacian Scores that also correspond to the highest F1-score.

\subsection{Synthetic Gaussian Mixtures}
We simulate $2$ modalities $\myX$ and $\myY$, where modality $\myX$ has $260$ samples with $130$ features and modality $\myY$ has $260$ samples with $90$ features. Both modalities have $3$ clusters in the data ($\myX$ has cluster $1$, $2$, $3$ and $\myY$ has cluster $1$, $2$, $4$, all labeled in red in Fig. \ref{fig:gauss_mat}), and each cluster has a set of informative features denoted as $\myvec{f}_{x,i}$ and $\myvec{f}_{x,i}$ ($i$ = $1$, $2$, $3$, $4$) with length $m_i$ ($i$ = $1$, $2$, $3$, $4$). Each set of these informative features is drawn from $N(\myvec{\mu_i},\myvec{I})$ independently for each sample, where $\myvec{\mu_i}$ is a vector of length $m_i$ drawn from $U(2,4)$ and $\myvec{I}$ is an $m_i \times m_i$ identity matrix.

By design, cluster $1$ and $2$ are shared between modalities with $m_1 = 20$ and $m_2 = 10$ in modality $\myX$, and $m_1 = 10$ and $m_2 = 10$ in modality $\myY$. On the other hand, cluster $3$ is specific to modality $\myX$ with $m_3 = 40$, and cluster $4$ is specific to modality $\myY$ with $m_4 = 40$. The remaining features are considered noisy features and are drawn from $N(0,1)$.

\subsection{Synthetic Developmental Tree}

We use \textit{generate\_data()} function from dyntoy \footnote{https://github.com/dynverse/dyntoy},a tree simulator package, to generate a dataset $\myvec{X}_0$ with $1000$ samples and $100$ features. Specifically, the parameter \textit{num\_branchpoints} is set to $1$, \textit{num\_cells} is set to $1000$, \textit{num\_features} is set to $100$, \textit{sample\_mean\_count} is set to $10$, \textit{sample\_dispersion\_count} is set to $50$, \textit{differentailly\_expressed\_rate} is set to $4$, and \textit{dropout\_probability\_factor} is set to $0$.

This step yields an initial data matrix $\myvec{X}_0 \in \mathbb{R}^{1000 \times 100}$, and these $1000$ samples are initially partitioned into $4$ groups: $G_1$ and $G_2$, $G_3$ and $G_4$, $G_5$, $G_6$ shown in Fig. \ref{fig:tree_umap}. For $\myvec{X}_0$, we further divide it into two halves, resulting in $2$ data matrices $\myX \in \mathbb{R}^{1000 \times 50}$ and $\myY \in \mathbb{R}^{1000 \times 50}$. We regard $\myX$ and $\myY$ as $2$ data modalities and these features as informative features contributing to the shared tree structure.

We further add $50$ features to each modality that are drawn from negative binomial distributions to construct the differential structures between modalities. Specifically, for modality $\myX$, the $50$ features of $G_1$ are drawn from $NB(\mu=4,\alpha=0.1)$ where $\mu$ and $\alpha$ are the mean and dispersion parameter of the negative binomial distribution, whereas the $50$ features of the other groups of samples are drawn from $NB(\mu=20,\alpha=0.1)$. Similarly, for modality $\myY$, the $50$ features of $G_3$ are drawn from $NB(\mu=4,\alpha=0.1)$ while the $50$ features of the other groups of samples are drawn from $NB(\mu=20,\alpha=0.1)$. Therefore, $G_1$ is bifurcated from $G_2$ and this structure is only observed in $\myX$, and $G_3$ is bifurcated from $G_4$ and this structure is only observed in $\myY$. 

Next, we row normalize each data matrix with a scaling factor $1e4$, and log1p transform the data. Then we standardize the features by z-scoring. At the end, we add $200$ features drawn from $N(0,1)$ to each modality as the noisy features.

\subsection{CITE-seq}

The human cord blood mononuclear cells (CBMCs) CITE-seq data was generated by \cite{stoeckius2017simultaneous}, where the expression levels of both RNA and protein are measured for the same cells. We analyze $3$ cell types: Erythroid cells, CD 34+ cells, and Murine cells. We row normalize each data matrix for both modalities. For the gene expression matrix (RNA), we filter the genes by standard deviation and keep the top $500$ variable genes. Then for both matrices, we standardize the features by z-scoring. 

\end{document}